\documentclass[10pt,twocolumn,letterpaper]{article}

\usepackage[pagenumbers]{cvpr} 

\usepackage{graphicx}
\usepackage{amsmath}
\usepackage{amssymb}
\usepackage{booktabs}

\usepackage{bm}
\usepackage{siunitx}
\usepackage{stfloats}
\usepackage{multirow}
\usepackage{pifont}
\usepackage[utf8x]{inputenc} 
\newcommand{\cmark}{\ding{51}}%
\newcommand{\xmark}{\ding{55}}%

\usepackage[pagebackref,breaklinks,colorlinks]{hyperref}

\usepackage[capitalize]{cleveref}
\crefname{section}{Sec.}{Secs.}
\Crefname{section}{Section}{Sections}
\Crefname{table}{Table}{Tables}
\crefname{table}{Tab.}{Tabs.}

\def\cvprPaperID{****} 
\def\confName{CVPR}
\def\confYear{2022}

\newcommand{\Tref}[1]{Table~\ref{#1}}
\newcommand{\Eref}[1]{Equation~\ref{#1}}
\newcommand{\Fref}[1]{Figure~\ref{#1}}

\makeatletter
\newcommand\smallfootnote{\@setfontsize\smallfootnote{7.5}{8.5}}
\makeatother

\begin{document}

\title{Mobile-Former: Bridging MobileNet and Transformer}

\author{
{Yinpeng Chen\textsuperscript{1} \qquad Xiyang Dai\textsuperscript{1} \qquad Dongdong Chen\textsuperscript{1} \qquad Mengchen Liu\textsuperscript{1} \qquad Xiaoyi Dong\textsuperscript{2}} \\ 
{\qquad Lu Yuan\textsuperscript{1} \qquad Zicheng Liu\textsuperscript{1}}\\
\\ 
\qquad \qquad \qquad \qquad \textsuperscript{1} Microsoft \qquad \qquad \qquad \qquad
\textsuperscript{2} University of Science and Technology of China\\

{\tt\small\{yiche,xidai,dochen,mengcliu,luyuan,zliu\}@microsoft.com}, \qquad
\tt\small{dlight@mail.ustc.edu.cn} \qquad

}
\maketitle

\begin{abstract}
    We present Mobile-Former, a parallel design of MobileNet and transformer with a two-way bridge in between. This structure leverages the advantages of MobileNet at local processing and transformer at global interaction. And the bridge enables bidirectional fusion of local and global features. Different from recent works on vision transformer, the transformer in Mobile-Former contains very few tokens (e.g. 6 or fewer tokens) that are randomly initialized to learn global priors, resulting in low computational cost. Combining with the proposed light-weight cross attention to model the bridge, Mobile-Former is not only computationally efficient, but also has more representation power. It outperforms MobileNetV3 at low FLOP regime from 25M to 500M FLOPs on ImageNet classification. For instance, Mobile-Former achieves 77.9\% top-1 accuracy at 294M FLOPs, gaining 1.3\% over MobileNetV3 but saving 17\% of computations. When transferring to object detection, Mobile-Former outperforms MobileNetV3 by 8.6 AP in RetinaNet framework. Furthermore, we build an efficient end-to-end detector by replacing backbone, encoder and decoder in DETR with Mobile-Former, which outperforms DETR by 1.1 AP but saves 52\% of computational cost and 36\% of parameters.
\end{abstract}

\vspace{1mm}
\section{Introduction}
\label{sec:intro}

\begin{figure}[t]
	\begin{center}
		\includegraphics[width=1.0\linewidth]{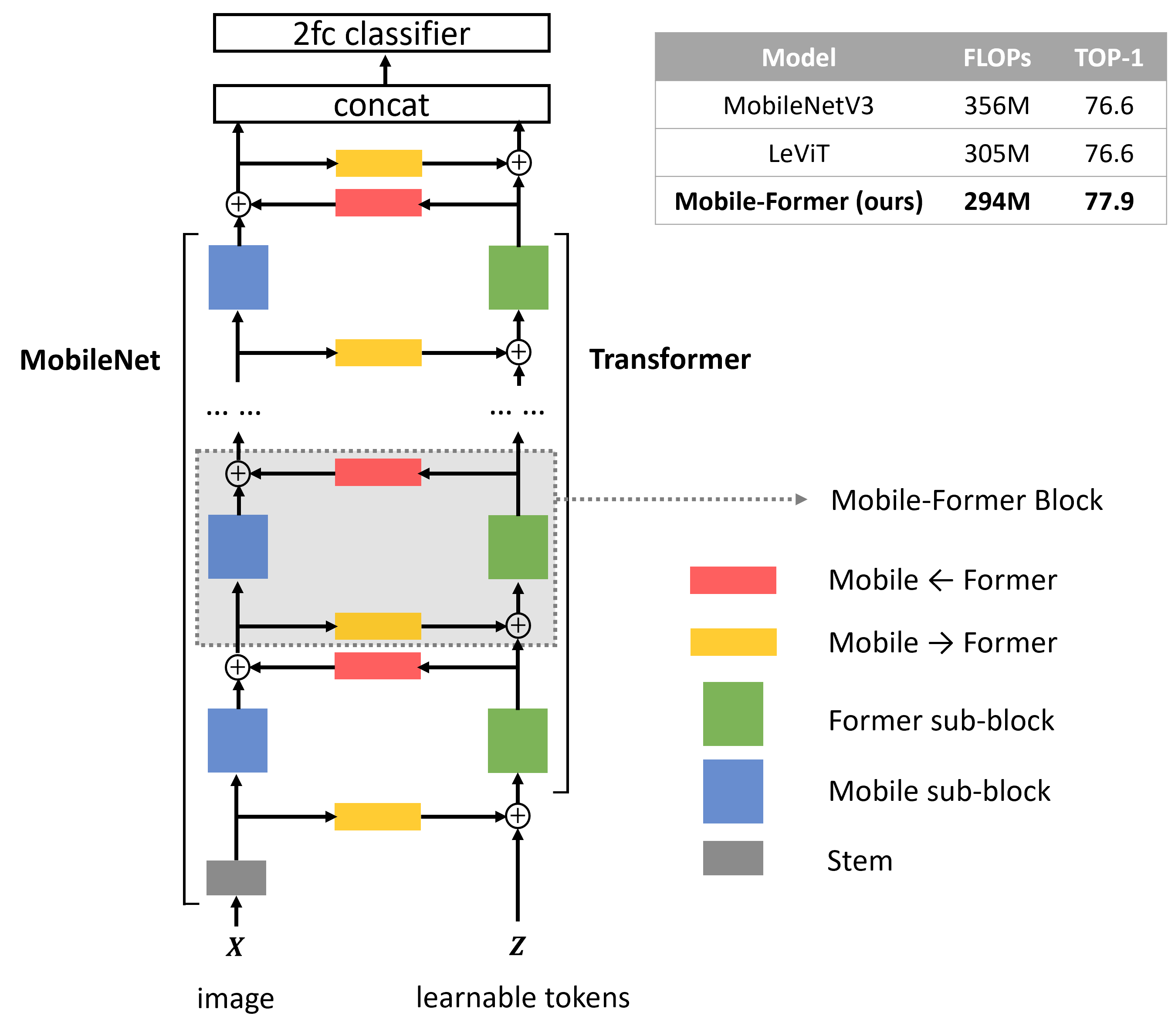}
	\end{center}
	\vspace{-4mm}
	\caption{\textbf{Overview of Mobile-Former}, which parallelizes MobileNet \cite{sandler2018mobilenetv2} on the left side and Transformer \cite{NIPS2017_transformer} on the right side. Different from vision transformer \cite{dosovitskiy2021vit} that uses image patches to form tokens, the transformer in Mobile-Former takes \textit{very few learnable tokens} as input that are randomly initialized. \textit{Mobile} (refers to MobileNet) and \textit{Former} (refers to transformer) communicate through a bidirectional bridge, which is modeled by the proposed light-weight cross attention. Best viewed in color.}
	\label{fig:overview}
	\vspace{-2mm}
\end{figure}

Recently, vision transformer (ViT) \cite{dosovitskiy2021vit, touvron2020deit} demonstrates the advantage of global processing and achieves significant performance boost over CNNs. However, when constraining the computational budget within 1G FLOPs, the gain of ViT diminishes. If we further challenge the computational cost, MobileNet \cite{howard2017mobilenets, sandler2018mobilenetv2, Howard_2019_ICCV_mbnetv3} and its extensions \cite{Han_2020_CVPR_ghostnet, li2021micronet} still dominate their backyard (e.g. fewer than 300M FLOPs for ImageNet classification) due to their efficiency in local processing filters via decomposition of depthwise and pointwise convolution. This in turn naturally raises a question: 
\begingroup
\addtolength\leftmargini{-0.05in}
\begin{quote}
     \textit{How to design \textbf{efficient} networks to \textbf{effectively} encode both local processing and global interaction?}
\end{quote}
\endgroup
\noindent A straightforward idea is to combine convolution and vision transformer. Recent works \cite{wu2021cvt, graham2021levit, Xiao-2021-early-cnns-help-transformers} show the benefit of  combining convolution and vision transformer in \textit{series}, either using convolution at the beginning or intertwining convolution into each transformer block.

In this paper, we shift the design paradigm from \textit{series} to \textit{parallel}, and propose a new network that parallelizes MobileNet and transformer with a two-way bridge in between (see \Fref{fig:overview}). We name it \textit{Mobile-Former}, where \textit{Mobile} refers to MobileNet and \textit{Former} stands for transformer.
\textit{Mobile} takes an image as input and stacks mobile (or inverted bottleneck) blocks \cite{sandler2018mobilenetv2}. It leverages the efficient depthwise and pointwise convolution to extract local features. \textit{Former} takes a few learnable tokens as input and stacks multi-head attention and feed-forward networks (FFN). These tokens are used to encode global features of the image.

\textit{Mobile} and \textit{Former} communicate through a two-way bridge to fuse local and global features. This is crucial since it feeds local features to \textit{Former's} tokens as well as introduces global views to every pixel of feature map in \textit{Mobile}. We propose a light-weight cross attention to model this bidirectional bridge by (a) performing the cross attention at the bottleneck of \textit{Mobile} where the number of channels is low, and (b) removing projections on query, key and value ($\bm{W}^Q$, $\bm{W}^K$, $\bm{W}^V$) from \textit{Mobile} side.  

This parallel structure with a bidirectional bridge leverages the advantages of both MobileNet and transformer. Decoupling of local and global features in parallel leverages MobileNet's efficiency in extracting local features as well as transformer's power in modeling global interaction. More importantly, this is achieved in an efficient way via a thin transformer with very few tokens and a light-weight bridge to exchange local and global features between \textit{Mobile} and \textit{Former}. The bridge and \textit{Former} consume less than 20\% of the total computational cost, but significantly improve the representation capability. This showcases an efficient and effective implementation of part-whole hierarchy \cite{Hinton2021-part-whole}.

\begin{figure}[t]
	\begin{center}
		\includegraphics[width=1.0\linewidth]{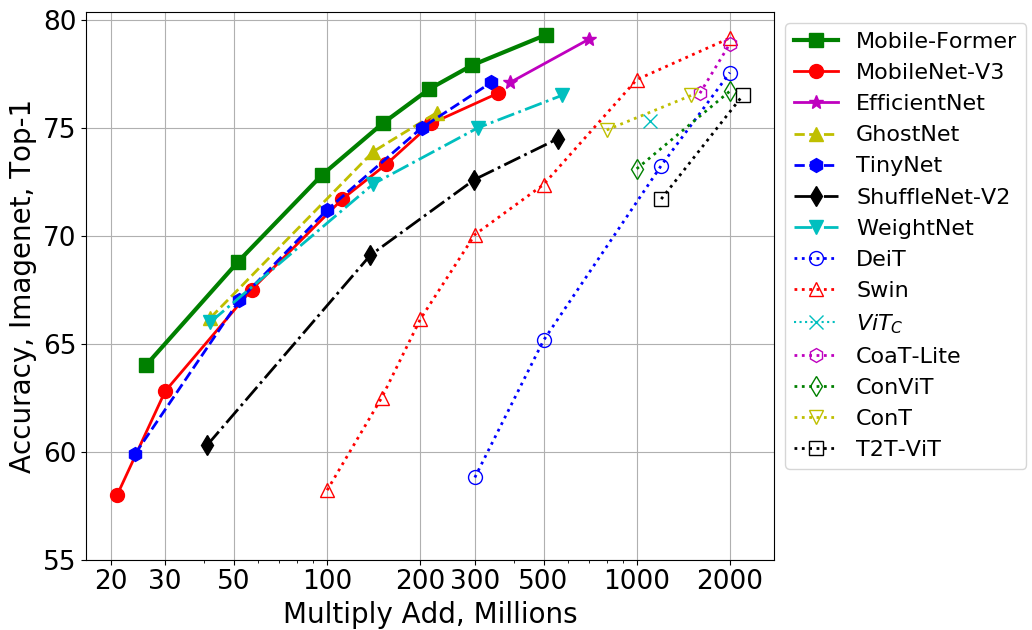}
	\end{center}
	\vspace{-3mm}
	\caption{\textbf{Comparison among Mobile-Former, efficient CNNs and vision transformers}, in terms of accuracy over FLOPs. The comparison is performed on ImageNet classification. Mobile-Former consistently outperforms both efficient CNNs and vision transformers in low FLOP regime (from 25M to 500M MAdds). Note that we implement Swin \cite{liu2021Swin} and DeiT \cite{touvron2020deit} at low computational budget from 100M to 2G FLOPs. Best viewed in color.}
	\label{fig:top-1}
	\vspace{-3mm}
\end{figure}

Mobile-Former achieves solid performance on both image classification and object detection. For example, it achieves 77.9\% top-1  accuracy on ImageNet classification at 294M FLOPs, outperforming MobileNetV3 \cite{Howard_2019_ICCV_mbnetv3} and LeViT \cite{graham2021levit} by a clear margin (see \Fref{fig:overview}). More importantly, Mobile-Former consistently outperforms both efficient CNNs and vision transformers from 25M to 500M FLOPs (see \Fref{fig:top-1}), showcasing the usage of transformer at the low FLOP regime where efficient CNNs dominate. 

When transferring from image classification to object detection, Mobile-Former significantly outperforms MobileNetV3 as backbone in RetinaNet \cite{Lin_2017_ICCV_retinanet_focal}, gaining 8.6 AP (35.8 vs. 27.2 AP) with even less computational cost. In addition, we build an efficient end-to-end detector by using Mobile-Former to replace backbone and encoder/decoder in DETR \cite{nicolas2020detr}. Using the same number of object queries (100), it gains 1.1 AP over DETR (43.1 vs. 42.0 AP) but has significantly fewer FLOPs (41G vs. 86G) and smaller model size (26.6M vs. 41.3M). 

Finally we note that \textit{exploring the optimal network parameters (e.g. width, height) in Mobile-Former is not a goal of this work}, rather we demonstrate that the parallel design provides an efficient and effective network architecture. 

\section{Related Work}
\label{sec:related-work}
\noindent \textbf{Light-weight convolutional neural networks (CNNs):} MobileNets \cite{howard2017mobilenets, sandler2018mobilenetv2, Howard_2019_ICCV_mbnetv3} efficiently encode local features by stacking depthwise and pointwise convolutions. 
ShuffleNet \cite{Zhang_2018_CVPR, ma_2018_ECCV} uses group convolution and channel shuffle to simplify pointwise convolution. MicroNet \cite{li2021micronet} presents micro-factorized convolution to handle extremely low FLOPs by lowering node connectivity to enlarge network width. Dynamic operators \cite{Hu_2018_CVPR, Yang2019CondConvCP, Chen2019DynamicCA, Chen2020DynamicReLU} have been studied to boost performance for MobileNet with negligible computational cost.
Other efficient operators include butterfly transform \cite{vahid_2020_CVPR}, cheap linear transformations in GhostNet \cite{Han_2020_CVPR_ghostnet}, and using additions to trade multiplications in AdderNet \cite{Chen_2020_CVPR_addernet}. In addition, 
MixConv \cite{Tan-bmvc2019-mixconv} explores mixing up multiple kernel sizes, and Sandglass \cite{Daquan_2020_ECCV_RethinkingBS} flips the structure of inverted residual block. EfficientNet \cite{tan-ICML19-efficientnet, Tan_2020_CVPR} and TinyNet \cite{NEURIPS2020_e069ea4c} study the compound scaling of depth, width and resolution. 

\vspace{1mm}
\noindent \textbf{Vision transformers (ViTs):} Recently, ViT \cite{dosovitskiy2021vit} and its follow-ups \cite{touvron2020deit, yuan2021tokens, liu2021Swin, dong2021cswin, Vaswani_2021_CVPR_halo} achieve impressive performance on multiple vision tasks. The original ViT requires training on large dataset such as JFT-300M. DeiT \cite{touvron2020deit} introduces training strategies on the smaller ImageNet-1K dataset. Later, hierarchical transformers are proposed to handle high resolution images. Swin \cite{liu2021Swin} computes self-attention within shifted local windows and CSWin \cite{dong2021cswin}  further improves it by introducing cross-shaped window. T2T-ViT \cite{yuan2021tokens} progressively converts an image to tokens by recursively aggregating neighboring tokens.
HaloNet \cite{Vaswani_2021_CVPR_halo} improves speed, memory usage and accuracy by two extensions (blocked local attention and attention downsampling).

\vspace{1mm}
\noindent \textbf{Combination of CNN and ViT:} Recent works \cite{Srinivas_2021_CVPR_bot, wu2021cvt, d2021convit, Xiao-2021-early-cnns-help-transformers, graham2021levit} show advantages of combining convolution and transformer. 
BoTNet \cite{Srinivas_2021_CVPR_bot} improves both instance segmentation and object detection by just using self-attention in the last three blocks of ResNet \cite{he2016deep}. ConViT \cite{d2021convit} presents a gated positional self-attention for soft convolutional inductive biases.
CvT \cite{wu2021cvt} introduces depthwise/pointwise convolution before multi-head attention. LeViT \cite{graham2021levit} and ViT$_C$ \cite{Xiao-2021-early-cnns-help-transformers} use convolutional stem 
to replace the patchify stem and achieve clear improvement. In this paper, we propose a different design that \textit{parallelizes} MobileNet and transformer with bidirectional cross attention in between. Our approach is efficient and effective, outperforming both efficient CNNs and ViT variants at low FLOP regime.

%

\section{Our Method: Mobile-Former}
\label{sec:method}
\begin{figure}[t]
	\begin{center}
		\includegraphics[width=0.8\linewidth]{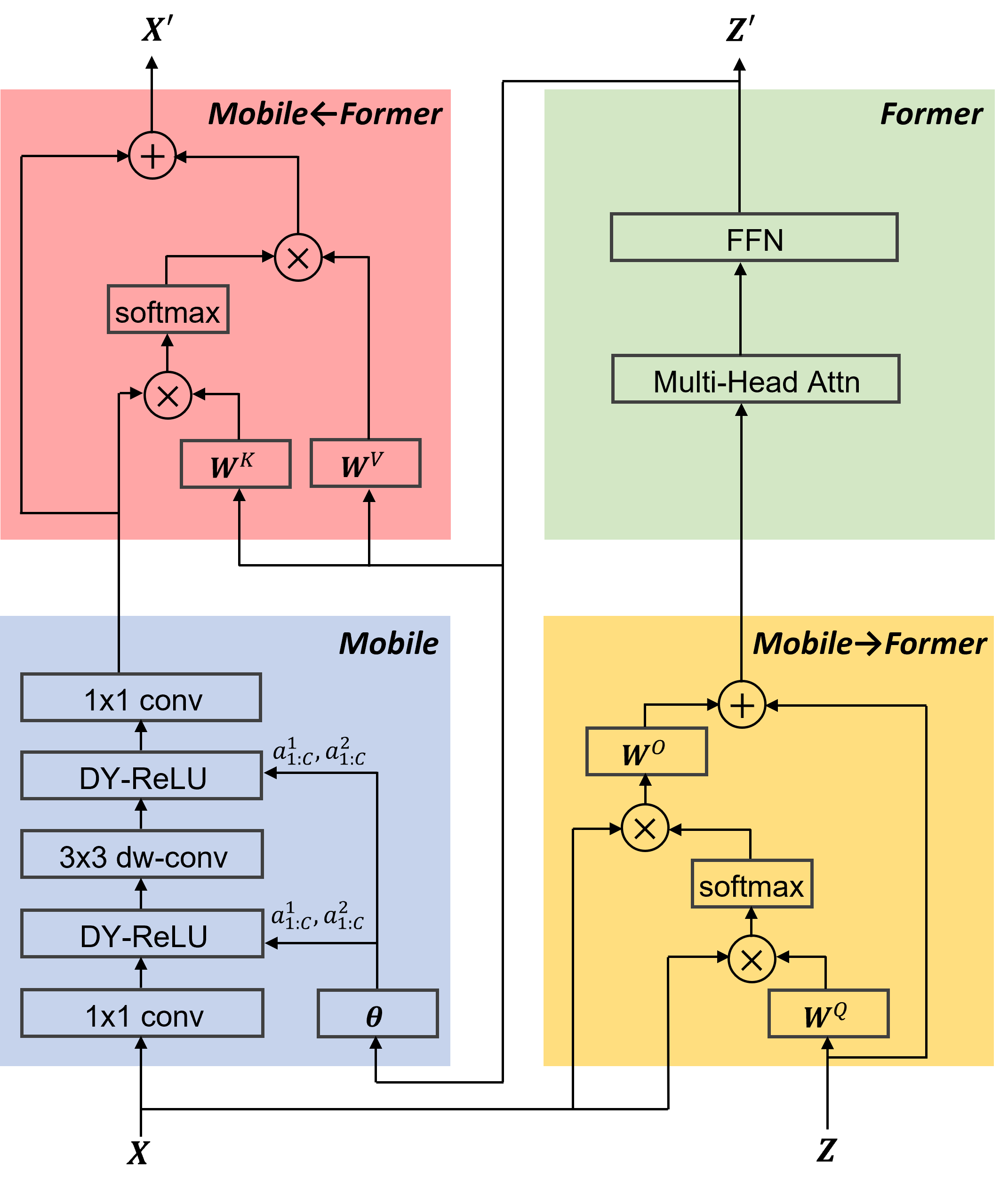}
	\end{center}
	\vspace{-4mm}
	\caption{\textbf{Mobile-Former block} that includes four modules: \textit{Mobile} sub-block modifies inverted bottleneck block in \cite{sandler2018mobilenetv2} by replacing ReLU with dynamic ReLU \cite{Chen2020DynamicReLU}. \textit{Mobile$\rightarrow$Former} uses light-weight cross attention to fuse local features into global features. \textit{Former} sub-block is a standard transformer block including multi-head attention and FFN. Note that the output of \textit{Former} is used to generate parameters for dynamic ReLU in \textit{Mobile} sub-block. \textit{Mobile$\leftarrow$Former} bridges from global to local features.}
	\label{fig:MF-block}
	\vspace{-3mm}
\end{figure}

In this section, we first overview the design of Mobile-Former (see \Fref{fig:overview}), and then discuss details within a Mobile-Former block (see \Fref{fig:MF-block}). Finally, we show the network specification and variants with different FLOPs.

\subsection{Overview}
\noindent \textbf{Parallel structure:}
Mobile-Former parallelizes MobileNet and transformer, and connects them by bidirectional cross attention (see \Fref{fig:overview}). 
\textit{Mobile} (refers to MobileNet) takes an image as input ($\bm{X} \in \mathbb{R}^{HW \times 3}$) and applies inverted bottleneck blocks \cite{sandler2018mobilenetv2} to extract local features. \textit{Former} (refers to transformer) takes learnable parameters (or tokens) as input, denoted as $\bm{Z} \in \mathbb{R}^{M \times d}$ where $M$ and $d$ are the number and dimension of tokens, respectively. These tokens are randomly initialized. Different from vision transformer (ViT) \cite{dosovitskiy2021vit}, where tokens project the local image patch linearly, \textit{Former} has significantly fewer tokens ($M \leq 6$ in this paper), each represents a global prior of the image. This results in much less computational cost. 

\vspace{1mm}
\noindent \textbf{Low cost two-way bridge:}
\textit{Mobile} and \textit{Former} communicate through a two-way bridge where local and global features are fused bidirectionally. The two directions are denoted as \textit{Mobile$\rightarrow$Former} and \textit{Mobile$\leftarrow$Former}, respectively. We propose a light-weight cross attention to model them, in which the projections ($\bm{W}^Q$, $\bm{W}^K$, $\bm{W}^V$) are removed from \textit{Mobile} side to save computations, but kept at \textit{Former} side. The cross attention is computed at the bottleneck of \textit{Mobile} where the number of channels is low. 
%
%
 %
 Specifically, the light-weight cross attention from local feature map $\bm{X}$ to global tokens $\bm{Z}$ is computed as:
\begin{align}
\mathcal{A}_{\bm{X}\rightarrow\bm{Z}}&=\left[Attn(\bm{\tilde{z}}_i\bm{W}_i^Q, \bm{\tilde{x}}_i, \bm{\tilde{x}}_i)\right]_{i=1:h}\bm{W}^O,
\label{eq:mobile2former}
\end{align}
where the local feature $\bm{X}$ and global tokens $\bm{Z}$ are split into $h$ heads as $\bm{X}=[\bm{\tilde{x}}_{1} \cdots \bm{\tilde{x}}_{h}]$, $\bm{Z}=[\bm{\tilde{z}}_{1} \cdots \bm{\tilde{z}}_{h}]$ for multi-head attention. The split for the $i^{th}$ head $\bm{\tilde{z}}_i \in \mathbb{R}^{M\times \frac{d}{h}}$ is different to the $i^{th}$ token $\bm{z}_i \in \mathbb{R}^{d}$. $\bm{W}_i^Q$ is the query projection matrix for the $i^{th}$ head. $\bm{W}^O$ is used to combine multiple heads together. $Attn(Q, K, V)$ is the standard attention function \cite{NIPS2017_transformer} over query $Q$, key $K$, and value $V$ as  $softmax(\frac{QK^T}{\sqrt{d_k}})V$. $[\cdot]_{1:h}$ denotes the concatenation of $h$ elements.
%
%
Note that the projection matrices for the key and value are removed from \textit{Mobile} side, while the project matrix $\bm{W}_i^Q$ for the query is kept at \textit{Former} side. 
%
Similarly, the cross attention from global to local is computed as:
\begin{align}
\mathcal{A}_{\bm{Z}\rightarrow\bm{X}}&=\left[ Attn(\bm{\tilde{x}}_i, \bm{\tilde{z}}_i\bm{W}_i^K, \bm{\tilde{z}}_i\bm{W}_i^V) \right]_{i=1:h},
\label{eq:Former2Mobile}
\end{align}
where $\bm{W}_i^K$ and $\bm{W}_i^V$ are the projection matrices for the key and value at \textit{Former} side. The projection matrix of the query is removed from \textit{Mobile} side. 

\subsection{Mobile-Former Block}
Mobile-Former consists of stacked \textit{Mobile-Former blocks} (see \Fref{fig:overview}). Each block has four pillars: a \textit{Mobile} sub-block, a \textit{Former} sub-block, and two-way cross attention \textit{Mobile$\leftarrow$Former} and \textit{Mobile$\rightarrow$Former} (shown in \Fref{fig:MF-block}). 

\vspace{2mm}
\noindent \textbf{Input and output:} 
Mobile-Former block has two inputs: (a) local feature map $\bm{X} \in \mathbb{R}^{HW \times C}$, which has $C$ channels over height $H$ and width $W$, and (b) global tokens $\bm{Z} \in \mathbb{R}^{M \times d}$, where $M$ and $d$ are the number and dimension of tokens, respectively. Note that $M$ and $d$ are identical across all blocks. Mobile-Former block outputs the updated local feature map $\bm{X}'$ and global tokens $\bm{Z}'$, which are used as input for the next block.  

\vspace{2mm}
\noindent \textbf{\textit{Mobile} sub-block:} As shown in \Fref{fig:MF-block}, \textit{Mobile} sub-block takes the feature map $\bm{X}$ as input and its output is taken as the input for \textit{Mobile$\leftarrow$Former}. It is slightly different to the inverted bottleneck block in \cite{sandler2018mobilenetv2} by replacing ReLU with dynamic ReLU \cite{Chen2020DynamicReLU} as the activation function. 
Different from the original dynamic ReLU, in which the parameters are generated by applying two MLP layers on the average pooled feature, we save the average pooling by applying the two MLP layers ($\bm{\theta}$ in \Fref{fig:MF-block}) on the first global token output $\bm{z}'_1$ from \textit{Former}. Note that the kernel size of depthwise convolution is 3$\times$3 for all blocks. 

\noindent \textbf{\textit{Former} sub-block:} \textit{Former} sub-block is a standard transformer block including a multi-head attention (MHA) and a feed-forward network (FFN). Expansion ratio 2 (instead of 4) is used in FFN. We follow \cite{NIPS2017_transformer} to use post layer normalization. \textit{Former} is processed between \textit{Mobile$\rightarrow$Former} and \textit{Mobile$\leftarrow$Former} (see \Fref{fig:MF-block}). 

 \vspace{2mm}
 \noindent \textbf{\textit{Mobile$\rightarrow$Former}:} The proposed light-weight cross attention (\Eref{eq:mobile2former}) is used to fuse local features $\bm{X}$ to global tokens $\bm{Z}$. Compared to the standard attention, the projection matrices for key $\bm{W}^K$ and value $\bm{W^V}$ (on the local features $\bm{X}$) are removed to save computations (see \Fref{fig:MF-block}). 

 \vspace{2mm}
 \noindent \textbf{\textit{Mobile$\leftarrow$Former}:} Here, the cross attention (\Eref{eq:Former2Mobile}) is on the opposite direction to \textit{Mobile$\rightarrow$Former}. It fuses global tokens to local features. The local features are the query and global tokens are the key and value. Therefore, we keep the projection matrices for the key $\bm{W}^K$ and value $\bm{W}^V$, but remove the projection matrix for the query $\bm{W}^Q$ to save computations, as shown in \Fref{fig:MF-block}. 
 
 \vspace{2mm}
 \noindent \textbf{Computational complexity:} The four pillars of a Mobile-Former block have different computational costs. Given an input feature map with size $HW \times C$, and $M$ global tokens with $d$ dimensions,  \textit{Mobile} consumes the most computations $O(HWC^2)$.
 \textit{Former} and the two-way bridge have light weight, consuming less than 20\% of the total computational cost. Specifically, \textit{Former} has complexity $O(M^2d+Md^2)$ for self attention and FFN. \textit{Mobile$\rightarrow$Former} and \textit{Mobile$\leftarrow$Former} share the complexity $O(MHWC+MdC)$ for cross attention.
 
\subsection{Network Specification}

\begin{table}[t!]
	\begin{center}
	    \footnotesize
	    \setlength{\tabcolsep}{1.7mm}{
		\begin{tabular}{c|c|c|c|c|c}
		    \specialrule{.1em}{.05em}{.05em} 
			Stage & Input & Operator & exp size & \#out & Stride   \\
		
			\specialrule{.1em}{.05em}{.05em} 
		    tokens & 6$\times$192  & -- & -- & -- &  --		 \\
			\hline
			stem & 224$^2\times$3  & conv2d, 3$\times$3 & -- & 16 &  2		 \\
			\hline
			1 & 112$^2\times$16  & bneck-lite & 32 & 16 & 1     \\
		    \hline
		    \multirow{2}{*}{2} & 112$^2\times$16  & Mobile-Former$^{\downarrow}$ & 96 & 24 & 2     \\
		     &56$^2\times$24  & Mobile-Former & 96 & 24 & 1     \\
		    \hline
		    \multirow{2}{*}{3} & 56$^2\times$24  & Mobile-Former$^{\downarrow}$ & 144 & 48 & 2     \\
		    & 28$^2\times$48  & Mobile-Former & 192 & 48 & 1     \\
		    \hline
		    
		    \multirow{4}{*}{4} & 28$^2\times$48  & Mobile-Former$^{\downarrow}$ & 288 & 96 & 2     \\
		    & 14$^2\times$96  & Mobile-Former & 384 & 96 & 1     \\
		    &14$^2\times$96  & Mobile-Former & 576 & 128 & 1     \\
		    &14$^2\times$128  & Mobile-Former & 768 & 128 & 1     \\
		    \hline
		    \multirow{4}{*}{5} &14$^2\times$128  & Mobile-Former$^{\downarrow}$ & 768 & 192 & 2     \\
		    & 7$^2\times$192  & Mobile-Former & 1152 & 192 & 1     \\
		    &7$^2\times$192  & Mobile-Former & 1152 & 192 & 1     \\
		    &7$^2\times$192  & conv2d, 1$\times$1 & -- & 1152 & 1     \\
		    \hline
		    \multirow{4}{*}{head} & 7$^2\times$1152  & pool, 7$\times$7 & -- & 1152 & --     \\
		    & 1$^2\times$1152  & concat w/ cls token & -- & 1344 & --     \\
		    &1$^2\times$1344  & FC & -- & 1920 & --     \\
		    &1$^2\times$1920  & FC & -- & 1000 & --     \\
			\specialrule{.1em}{.05em}{.05em} 
		\end{tabular}
		}
	\end{center}
	\vspace{-3mm}
	\caption{\textbf{Specification for Mobile-Former-294M}. ``bneck-lite" denotes the lite bottleneck block \cite{li2021micronet}. ``Mobile-Former$^{\downarrow}$" denotes the variant of downsample block.}
	\label{table:mobile-former-specification}
	\vspace{-3mm}
\end{table}

\noindent \textbf{Architecture:} 
\Tref{table:mobile-former-specification} shows a Mobile-Former architecture with 294M FLOPs for image size 224$\times$224, which stacks 11 Mobile-Former blocks at different input resolutions. All blocks have six global tokens with dimension 192. It starts with a 3$\times$3 convolution as stem and a lite bottleneck block \cite{li2021micronet} at stage 1, which expands and then squeezes the number of channels by stacking a 3$\times$3 depthwise and a pointwise convolution. Stage 2--5 consists of Mobile-Former blocks. Each stage handles downsampling by a downsample variant denoted as Mobile-Former$^{\downarrow}$ (see appendix~\ref{apx:arch-all} for details). The classification head applies average pooling on the local features, concatenates with the first global token, and then passes through two fully connected layers with h-swish \cite{Howard_2019_ICCV_mbnetv3} in between.


\vspace{1mm}
\noindent \textbf{Mobile-Former variants:} Mobile-Former has seven models with different computational costs from 26M to 508M FLOPs. They share the similar architecture, but have different width and height. We follow \cite{Xiao-2021-early-cnns-help-transformers} to refer our models by their FLOPs, e.g. Mobile-Former-294M, Mobile-Former-96M. The details of network architecture for these models are listed in 
appendix~\ref{apx:arch-all} (see \Tref{table:mf-achs}).

\section{Efficient End-to-End Object Detection}
\label{sec:method}
Mobile-Former can be easily used for object detection in both backbone and head, providing an \textit{efficient end-to-end} detector. Using the same number of object queries (100), it outperforms DETR \cite{nicolas2020detr}, but uses much lower FLOPs. 

\begin{figure}[t]
	\begin{center}
		\includegraphics[width=1.0\linewidth]{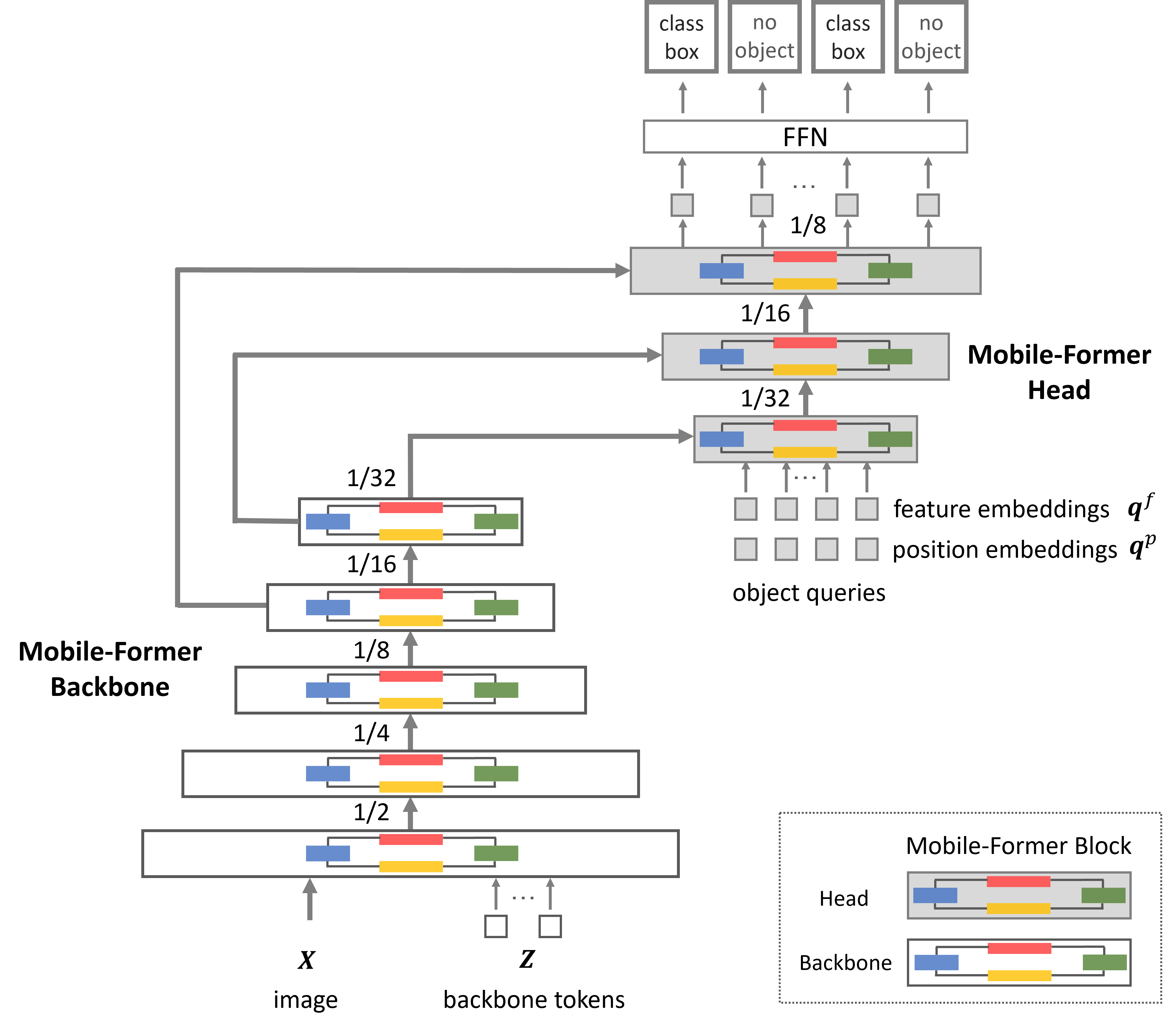}
	\end{center}
	\vspace{-4mm}
	\caption{\textbf{Mobile-Former for object detection}. Both backbone and head use Mobile-Former blocks (see \Fref{fig:overview}, \ref{fig:MF-block}). The backbone has 6 global tokens while the head has 100 object queries. All object queries pass through multiple resolutions ($\frac{1}{32}$, $\frac{1}{16}$, $\frac{1}{8}$) in the head. Similar to DETR \cite{nicolas2020detr}, feed forward network (FFN) is used to predict class label and bounding box.  Best viewed in color.}
	\label{fig:obj-det}
	\vspace{-2mm}
\end{figure}

\vspace{2mm}
\noindent \textbf{Backbone--Head architecture:}
We use Mobile-Former blocks in both backbone and head (see Figure \ref{fig:obj-det}), which
have separate tokens. The backbone has six global tokens while the head has 100 object queries generated similarly to DETR \cite{nicolas2020detr}. Different from DETR that has a single scale ($\frac{1}{32}$ or $\frac{1}{16}$) in the head, Mobile-Former head employs multi-scales ($\frac{1}{32}$, $\frac{1}{16}$, $\frac{1}{8}$) in low FLOPs due to its computational efficiency. The upsampling is achieved via bilinear interpolation followed by adding the feature output from the backbone (with the same resolution). 
All object queries progressively refine their representation across scales from coarse to fine,
saving the manual process in FPN \cite{Lin_FPN} to allocate objects across scales by size. We follow DETR to use prediction FFNs and auxiliary losses in the head during the training.
The head is trained from scratch and the backbone is pretrained on ImageNet. 
Our end-to-end Mobile-Former detector is computationally efficient. The total cost of E2E-MF-508M that uses Mobile-Former-508M as backbone and nine Mobile-Former blocks in the head is 41.4G FLOPs, significantly less than DETR (86G FLOPs). But it outperforms DETR by 1.1 AP (43.1 vs. 42.0 AP). The details of head structure is listed in appendix \ref{apx:arch-all} (see \Tref{table:e2e-head-achs}).

\vspace{2mm}
\noindent \textbf{Spatial-aware dynamic ReLU in backbone:}
We extend dynamic ReLU in the backbone from \textit{spatial-shared} to \textit{spatial-aware} by involving all global tokens to generate parameters, rather than just using the first one, as these tokens have different spatial focuses. Let us denote the parameter generation of the spatial-shared dynamic ReLU as $\bm{\theta}=f(\bm{z}_1)$, where $\bm{z}_1$ is the first global token and $f(\cdot)$ is modeled by two MLP layers with ReLU in between. By contrast, the spatial-aware dynamic ReLU generates parameters $\bm{\theta}_i$ per spatial position $i$ in a feature map, by using all global tokens $\{\bm{z}_j\}$ as follows:
\begin{align}
\bm{\theta}_i = \sum_j \alpha_{i, j} f(\bm{z}_j), s.t. \sum_j \alpha_{i, j}=1,
\label{eq:spa-dy-relu}
\end{align}
where $\alpha_{i, j}$ is the attention between the feature at position $i$ and token $\bm{z}_j$. Its calculation is cheap by just normalizing the cross attention obtained in \textit{Mobile$\rightarrow$Former} along tokens $\{\bm{z}_j\}$. Spatial-aware dynamic ReLU is on par with its spatial-shared counterpart on image classification, but gains 1.1 AP on COCO object detection (see \Tref{table:coco-det-detr-ablation}).

\vspace{2mm}
\noindent \textbf{Adapting position embedding in head:}
Different from DETR \cite{nicolas2020detr} that shared the position embedding of object queries across all decoder layers, we refine the position embedding after each block in the head as the feature map changes per block. Let us denote the feature and position embedding of a query at the $k^{th}$ block as $\bm{q}^f_k$ and $\bm{q}^p_k$, respectively. The sum of them ($\bm{q}^f_k+\bm{q}^p_k$) is used to compute cross attention between object queries and feature map as well as self attention among object queries, after which the feature embedding is updated as the input for the next block $\bm{q}^f_{k+1}$. Here, we adapt the position embedding based upon the feature embedding as:
\begin{align}
\bm{q}^p_{k+1} = \bm{q}^p_{k} + g(\bm{q}^f_{k+1}),
\label{eq:query-refine}
\end{align}
where the adaptation function $g(\cdot)$ is implemented by two MLP layers with ReLU in between. Thus, object queries can adapt their positions across blocks based on the content.

\section{Experimental Results}
We evaluate the proposed Mobile-Former on both ImageNet classification \cite{deng2009imagenet}, and COCO object detection \cite{lin2014microsoft}. 

\subsection{ImageNet Classification}
Image classification experiments are conducted on ImageNet \cite{deng2009imagenet} that has 1000 classes, including 1,281,167 images for training and 50,000 images for validation.

\vspace{2mm}
\noindent \textbf{Training setup:} The image resolution is 224$\times$224. All models are trained from scratch using AdamW \cite{loshchilov2018decoupled} optimizer for 450 epochs with cosine learning rate decay. A batch size of 1024 is used. Data augmentation includes Mixup \cite{zhang2018mixup}, auto-augmentation \cite{Cubuk_2019_CVPR}, and random erasing \cite{zhong2020random}. Different combinations of initial learning rate, weight decay and dropout are used for models with different complexities, which are listed in 
appendix \ref{apx:arch-all} (see \Tref{table:imagenet-hyper}).

\vspace{1mm}
\noindent \textbf{Comparison with efficient CNNs:} 
\Tref{table:imagenet-cls-cnn} shows the comparison between Mobile-Former and classic efficient CNNs: (a) MobileNetV3 \cite{Howard_2019_ICCV_mbnetv3}, (b) EfficientNet \cite{tan-ICML19-efficientnet}, and (c) ShuffleNetV2 \cite{ma_2018_ECCV} and its extension WeightNet \cite{Ma_2020_eccv_WeightNetRT}. The comparison covers the FLOP range from 26M to 508M, organized in seven groups based on similar FLOPs. Mobile-Former consistently outperforms efficient CNNs with even less computational cost except the group around 150M FLOPs, where Mobile-Former costs slightly more FLOPs than ShuffleNet/WeightNet (151M vs. 138M/141M), but achieves significantly higher top-1 accuracy (75.2\% vs. 69.1\%/72.4\%). This demonstrates that our parallel design improves the representation capability efficiently.

\begin{table}[t!]
	\begin{center}
	    \footnotesize
	    \setlength{\tabcolsep}{0.7mm}{
		\begin{tabular}{l|crr|c}
		    \specialrule{.1em}{.05em}{.05em} 
			Model & Input & \#Params & MAdds &Top-1 \\
		
			\specialrule{.1em}{.05em}{.05em} 
			MobileNetV3 Small 1.0$\times$ \cite{Howard_2019_ICCV_mbnetv3} & 160$^2$ & 2.5M & 30M  & 62.8  		 \\
			\textbf{Mobile-Former-26M}  & 224$^2$ & 3.2M & \textbf{26M} & \textbf{64.0} \\
			\hline
			MobileNetV3 Small 1.0$\times$ \cite{Howard_2019_ICCV_mbnetv3} & 224$^2$ & 2.5M & 57M  & 67.5 		 \\
			\textbf{Mobile-Former-52M} & 224$^2$ & 3.5M &\textbf{52M} & \textbf{68.7} \\
			\hline
			MobileNetV3 1.0$\times$ \cite{Howard_2019_ICCV_mbnetv3} & 160$^2$ & 5.4M & 112M  & 71.7 		 \\
			\textbf{Mobile-Former-96M} & 224$^2$ & 4.6M & \textbf{96M} & \textbf{72.8} \\
			\hline
			ShuffleNetV2 1.0$\times$ \cite{ma_2018_ECCV} & 224$^2$ & 2.2M & \textbf{138M} & 69.1  \\
			ShuffleNetV2 1.0$\times$+WeightNet 4$\times$ \cite{Ma_2020_eccv_WeightNetRT} & 224$^2$ & 5.1M & 141M & 72.4  \\
			MobileNetV3 0.75$\times$ \cite{Howard_2019_ICCV_mbnetv3} & 224$^2$ & 4.0M & 155M  & 73.3  		 \\
			\textbf{Mobile-Former-151M} & 224$^2$ & 7.6M & 151M & \textbf{75.2} \\
			\hline
			MobileNetV3 1.0$\times$ \cite{Howard_2019_ICCV_mbnetv3} & 224$^2$ & 5.4M & 217M  & 75.2		 \\
			\textbf{Mobile-Former-214M}  & 224$^2$ & 9.4M & \textbf{214M} & \textbf{76.7} \\
			
			\hline
		    ShuffleNetV2 1.5$\times$ \cite{ma_2018_ECCV}   & 224$^2$ & 3.5M & 299M & 72.6 \\
			ShuffleNetV2 1.5$\times$+WeightNet 4$\times$ \cite{Ma_2020_eccv_WeightNetRT}   & 224$^2$ & 9.6M & 307M & 75.0 \\
			MobileNetV3 1.25$\times$ \cite{Howard_2019_ICCV_mbnetv3}   & 224$^2$ & 7.5M & 356M  & 76.6 		 \\
			EfficientNet-B0 \cite{tan-ICML19-efficientnet}  & 224$^2$ & 5.3M & 390M & 77.1 \\
			\textbf{Mobile-Former-294M} & 224$^2$ & 11.4M & \textbf{294M} & \textbf{77.9} \\
			\hline
			ShuffleNetV2 2$\times$ \cite{ma_2018_ECCV}  & 224$^2$ & 5.5M & 557M & 74.5  \\
			ShuffleNetV2 2$\times$+WeightNet 4$\times$ \cite{Ma_2020_eccv_WeightNetRT} & 224$^2$ & 18.1M & 573M & 76.5  \\
			\textbf{Mobile-Former-508M} & 224$^2$ & 14.0M & \textbf{508M} & \textbf{79.3} \\
			\specialrule{.1em}{.05em}{.05em} 
		\end{tabular}
		}
	\end{center}
	\vspace{-3mm}
	\caption{\textbf{Comparing Mobile-Former with efficient CNNs} evaluated on ImageNet \cite{deng2009imagenet} classification. 
	}
	\label{table:imagenet-cls-cnn}
	\vspace{-1mm}
\end{table}

\begin{table}[t!]
	\begin{center}
	    \footnotesize
	    \setlength{\tabcolsep}{2.6mm}{
		\begin{tabular}{l|crr|c}
		    \specialrule{.1em}{.05em}{.05em} 
			Model & Input & \#Params & MAdds &Top-1 \\
		
			\specialrule{.1em}{.05em}{.05em} 
			T2T-ViT-7 \cite{yuan2021tokens} & 224$^2$ & 4.3M & 1.2G & 71.7 \\
			DeiT-Tiny \cite{touvron2020deit} & 224$^2$ & 5.7M & 1.2G & 72.2 \\
			ConViT-Tiny \cite{d2021convit} & 224$^2$ & 6.0M & 1.0G & 73.1 \\
			ConT-Ti \cite{yan2104contnet} & 224$^2$ & 5.8M & 0.8G & 74.9 \\
			ViT$_C$ \cite{Xiao-2021-early-cnns-help-transformers}  & 224$^2$ & 4.6M & 1.1G & 75.3 \\
			ConT-S \cite{yan2104contnet} & 224$^2$ & 10.1M & 1.5G & 76.5 \\
			Swin-1G \cite{liu2021Swin} $^{\ddag}$ &  224$^2$ & 7.3M & 1.0G & 77.3  \\
			\textbf{Mobile-Former-294M} & 224$^2$ & 11.4M & \textbf{294M} & \textbf{77.9} \\
			\hline

			PVT-Tiny \cite{wang2021pvtv1} & 224$^2$ & 13.2M & 1.9G & 75.1 \\
			T2T-ViT-12 \cite{yuan2021tokens} & 224$^2$ & 6.9M & 2.2G & 76.5 \\
			
			CoaT-Lite Tiny \cite{xu2021coscale} & 224$^2$ & 5.7M & 1.6G & 76.6 \\
			ConViT-Tiny+ \cite{d2021convit} & 224$^2$ & 10.0M & 2G & 76.7 \\
			DeiT-2G \cite{touvron2020deit} $^{\ddag}$ & 224$^2$ & 9.5M & 2.0G & 77.6 \\
			CoaT-Lite Mini \cite{xu2021coscale} & 224$^2$ & 11.0M & 2.0G & 78.9 \\
			BoT-S1-50 \cite{Srinivas_2021_CVPR_bot} & 224$^2$ & 20.8M & 4.3G & 79.1 \\
			Swin-2G \cite{liu2021Swin} $^{\ddag}$ &  224$^2$ & 12.8M & 2.0G & 79.2  \\
			\textbf{Mobile-Former-508M} & 224$^2$ & 14.0M & \textbf{508M} & \textbf{79.3} \\
			\specialrule{.1em}{.05em}{.05em} 
		\end{tabular}
		}
	\end{center}
	\vspace{-3mm}
	\caption{\textbf{Comparing Mobile-Former with vision transformer variants} evaluated on ImageNet \cite{deng2009imagenet} classification. Here, we choose ViT variants that use image resolution 224$\times$224 and are trained \textit{without} distillation from a teacher network. We group ViT models based on FLOPs (using 1.5G as threshold) and rank them based on top-1 accuracy. $^{\ddag}$ indicates our implementation.
	}
	\label{table:imagenet-cls-vit}
	\vspace{-1mm}
\end{table}

\vspace{1mm}
\noindent \textbf{Comparison with ViTs:} 
In \Tref{table:imagenet-cls-vit}, we compare Mobile-Former with multiple variants (DeiT \cite{touvron2020deit}, T2T-ViT \cite{yuan2021tokens}, PVT \cite{wang2021pvtv1}, ConViT \cite{d2021convit}, CoaT \cite{xu2021coscale}, ViT$_C$ \cite{Xiao-2021-early-cnns-help-transformers}, Swin \cite{liu2021Swin}) of vision transformer. All variants use image resolution 224$\times$224 and are trained \textit{without} distillation from a teacher network. Mobile-Former achieves higher accuracy but uses 3$\sim$4 times less computational cost. This is because that Mobile-Former uses significantly fewer tokens to model global interaction and leverages MobileNet to extract local features efficiently. 
Note that our Mobile-Former (trained in 450 epochs without distillation) even outperforms LeViT \cite{graham2021levit} which leverages the distillation of a teacher network and much longer training (1000 epochs). Our method achieves higher top-1 accuracy (77.9\% vs. 76.6\%) but uses fewer FLOPs (294M vs. 305M) than LeViT. 

\vspace{1mm}
\noindent \textbf{Accuracy--FLOP plot:}
\Fref{fig:top-1} compares Mobile-Former with more CNN models (e.g. GhostNet \cite{Han_2020_CVPR_ghostnet}) and vision transformer variants (e.g. Swin \cite{liu2021Swin} and DeiT \cite{touvron2020deit}) in one plot. We implement Swin and DeiT from 100M to 2G FLOPs, by carefully reducing network width and height. Mobile-Former clearly outperforms both CNNs and ViT variants, showcasing the parallel design to integrate MobileNet and transformer. Although vision transformers are inferior to efficient CNNs by a large margin, our work demonstrates that the transformer can also contribute to the low FLOP regime with proper architecture design.   

\subsection{Ablations}
In this subsection, we show Mobile-Former is effective and efficient via several ablations performed on ImageNet classification. Here, Mobile-Former-294M is used and all models are trained for 300 epochs. Moreover, we summarize interesting observations on the visualization of the two-way bridge between \textit{Mobile} and \textit{Former}. 

\vspace{2mm}
\noindent \textbf{Mobile-Former is effective:}
\begin{table}[t!]
	\begin{center}
	    \footnotesize
	    \setlength{\tabcolsep}{1.5mm}{
		\begin{tabular}{l|cc|ll}
		    \specialrule{.1em}{.05em}{.05em} 
			Model &  \#Params & MAdds &Top-1 &	Top-5   \\
		
			\specialrule{.1em}{.05em}{.05em} 
			\textit{Mobile} (using ReLU)  & 6.1M & 259M & 74.2 &  91.8		 \\
			+ \textit{Former} and Bridge  & 10.1M & 290M & 76.8$_{(+2.6)}$ &  93.2$_{(+1.4)}$		 \\
			+ DY-ReLU in \textit{Mobile}  & 11.4M & 294M & 77.8$_{(+1.0)}$ &  93.7$_{(+0.5)}$		 \\
			\specialrule{.1em}{.05em}{.05em} 
		\end{tabular}
		}
	\end{center}
	\vspace{-3mm}
	\caption{\textbf{Ablation of \textit{Former}+bridge and dynamic ReLU} evaluated on ImageNet classification. Mobile-Former-294M is used.}
	\label{table:ablation-former-bridge}
	\vspace{-1mm}
\end{table}
Mobile-Former is more effective than MobileNet as it encodes global interaction via \textit{Former}, resulting in more accurate prediction. As shown in \Tref{table:ablation-former-bridge}, adding \textit{Former} and bridge (\textit{Mobile$\rightarrow$Former} and \textit{Mobile$\leftarrow$Former}) only costs 12\% of the computational cost, but gains 2.6\% top-1 accuracy over the baseline that uses \textit{Mobile} alone. In addition, using dynamic ReLU \cite{Chen2020DynamicReLU} in \textit{Mobile} sub-block (see \Fref{fig:MF-block}) gains additional 1.0\% top-1 accuracy. Note that the parameters in dynamic ReLU is generated by using the first global token.
This validates our parallel design in Mobile-Former. We also find that increasing kernel size (3$\times$3 $\rightarrow$ 5$\times$5) of the depthwise convolution in \textit{Mobile} only introduces negligible gain 
(see \Tref{table:ablation-kernel-size} in appendix~\ref{apx:more-results}), as the reception field of \textit{Mobile} is enlarged by fusing global features from \textit{Former}.

%

\vspace{2mm}
\noindent \textbf{Mobile-Former is efficient:}
Mobile-Former is not only effective in encoding both local processing and global interaction, but achieves this \textit{efficiently}. Ablations below show that \textit{Former only requires a few global tokens with low dimension}. In addition, 
the efficient parallel design of Mobile-Former is stable when removing FFN in \textit{Former} or replacing multi-head attention with position mixing MLP \cite{tolstikhin2021mlpmixer}.

\vspace{1mm}
\noindent \textit{Number of tokens in \textit{Former}:} \Tref{table:ablation-number-of-tokens} shows the ImageNet classification results for using different number of global tokens in \textit{Former}. The token dimension is 192. Interestingly, even a single global token achieves a good performance (77.1\% top-1 accuracy).
Additional improvement (0.5\% and 0.7\% top-1 accuracy) is achieved when using 3 and 6 tokens. The improvement stops when more than 6 tokens are used. Such compactness of global tokens is a key contributor to the efficiency of Mobile-Former.
\begin{table}[t!]
	\begin{center}
	    \footnotesize
	    \setlength{\tabcolsep}{4.2mm}{
		\begin{tabular}{c|cc|cc}
		    \specialrule{.1em}{.05em}{.05em} 
			\#Tokens & \#Params & MAdds &Top-1 &	Top-5   \\
		
			\specialrule{.1em}{.05em}{.05em} 
			1  & 11.4M & 269M & 77.1 &  93.2		 \\
			3  & 11.4M & 279M & 77.6 &  93.6		 \\
			6  & 11.4M & 294M & 77.8 &  93.7		 \\
			9  & 11.4M & 309M & 77.7 & 93.8   		 \\
			\specialrule{.1em}{.05em}{.05em} 
		\end{tabular}
		}
	\end{center}
	\vspace{-3mm}
	\caption{\textbf{Ablation of the number of tokens} on ImageNet classification. Mobile-Former-294M is used.}
	\label{table:ablation-number-of-tokens}
	\vspace{-2mm}
\end{table}

\vspace{1mm}
\noindent \textit{Token dimension:} \Tref{table:ablation-tokens-dimension} shows the results for different token lengths (or dimensions). Here, six global tokens are used in \textit{Former}. The accuracy improves from 76.8\% to 77.8\% when token dimension increases from 64 to 192, but converges when higher dimension is used. This further supports the efficiency of \textit{Former}. With six tokens of dimension 192, the total computational cost of \textit{Former} and the bridge only consumes 12\% of the overall budget (35M/294M).

\begin{table}[t!]
	\begin{center}
	    \footnotesize
	    \setlength{\tabcolsep}{3.1mm}{
		\begin{tabular}{c|cc|cc}
		    \specialrule{.1em}{.05em}{.05em} 
			Token Dimension & \#Params & MAdds &Top-1 &	Top-5   \\
		
			\specialrule{.1em}{.05em}{.05em} 
			64  & 7.3M & 277M & 76.8 &  93.1		 \\
			128  & 9.1M & 284M & 77.3 &  93.5		 \\
			192  & 11.4M & 294M & 77.8 &  93.7		 \\
			256  & 14.3M & 308M & 77.8 &  93.7		 \\
			320  & 17.9M & 325M & 77.6 &  93.6		 \\
			\specialrule{.1em}{.05em}{.05em} 
		\end{tabular}
		}
	\end{center}
	\vspace{-3mm}
	\caption{\textbf{Ablation of token dimension} on ImageNet classification. Mobile-Former-294M is used.}
	\label{table:ablation-tokens-dimension}
	\vspace{-1mm}
\end{table}

\vspace{1mm}
\noindent \textit{FFN in \textit{Former}:}
As shown in \Tref{table:ablation-ffn-mlp}, removing FFN introduces a small drop in top-1 accuracy ($-$0.3\%). Compared to the important role of FFN in the original vision transformer, FFN has limited contribution in Mobile-Former. We believe this is because FFN is not the only module for channel fusion in Mobile-Former. The 1$\times$1 convolution in \textit{Mobile} helps the channel fusion of local features, while the projection matrix $\bm{W}^O$ in Mobile$\rightarrow$Former (see \Eref{eq:mobile2former}) contributes to the fusion between local and global features.

\begin{table}[t!]
	\begin{center}
	    \footnotesize
	    \setlength{\tabcolsep}{2.5mm}{
		\begin{tabular}{cc|cc|cc}
		    \specialrule{.1em}{.05em}{.05em} 
			 Attention & FFN & \#Params & MAdds &Top-1 &	Top-5   \\
		
			\specialrule{.1em}{.05em}{.05em}
			MHA & \cmark  & 11.4M & 294M & 77.8 &  93.7		 \\
			MHA & \xmark & 9.8M & 284M & 77.5 & 	93.6	 \\
			Pos-Mix-MLP & \cmark  & 10.5M & 284M & 77.3 &  93.5		 \\

			\specialrule{.1em}{.05em}{.05em} 
		\end{tabular}
		}
	\end{center}
	\vspace{-3mm}
	\caption{\textbf{Ablation of multi-head attention (MHA) and FFN} on ImageNet classification. Mobile-Former-294M is used.}
	\label{table:ablation-ffn-mlp}
	\vspace{-1mm}
\end{table}

\vspace{1mm}
\noindent \textit{Multi-head attention (MHA) vs. position-mixing MLP:} \Tref{table:ablation-ffn-mlp} shows the results of replacing multi-head attention (MHA) with token/position mixing MLP \cite{tolstikhin2021mlpmixer} in both \textit{Former} and bridge (\textit{Mobile$\rightarrow$Former} and \textit{Mobile$\leftarrow$Former}). The top-1 accuracy drops from 77.8\% to 77.3\%. The implementation of MLP is more efficient by a single matrix multiplication, but it is static (i.e. not adaptive to different input images).

\vspace{1mm}
\noindent \textbf{Mobile-Former is explainable:}
We observe three interesting patterns in the two-way bridge (\textit{Mobile$\rightarrow$Former} and \textit{Mobile$\leftarrow$Former}). First, from low to high levels, the global tokens change their focus from edges/corners, to foreground/background, and finally on the most discriminative region. Second, the cross attention has more diversity across tokens at lower levels than high levels. Thirdly, the separation between foreground and background is found at middle layers of \textit{Mobile$\leftarrow$Former}. The detailed visualization is shown in appendix~\ref{apx:vis}.

\subsection{Object Detection}

\begin{table}[t!]
	\begin{center}
	    \smallfootnote
	    \setlength{\tabcolsep}{0.8mm}{
		\begin{tabular}{l|ccc|ccc|c|c}
		    \specialrule{.1em}{.05em}{.05em} 
			\multirow{2}{*}{Model} & \multirow{2}{*}{AP} & \multirow{2}{*}{AP\textsubscript{50}} & \multirow{2}{*}{AP\textsubscript{75}} & \multirow{2}{*}{AP\textsubscript{S}} & \multirow{2}{*}{AP\textsubscript{M}} & \multirow{2}{*}{AP\textsubscript{L}} & MAdds & \#Params  \\
	         &  &  &  &  &  &  & (G) & (M)  \\	
			\specialrule{.1em}{.05em}{.05em} 
Shuffle-V2 \cite{ma_2018_ECCV} & 25.9 & 41.9 & 26.9 & 12.4 & 28.0 & 36.4 & \textbf{2.6} (161) & 0.8 (10.4)\\
     \textbf{MF-151M} &  \textbf{34.2}  & 53.4 & 36.0 & 19.9 & 36.8 & 45.3 & \textbf{2.6} (161) & 4.9 (14.4)\\ 
		    \hline
Mobile-V3 \cite{Howard_2019_ICCV_mbnetv3} &  27.2 & 43.9 & 28.3 & 13.5 & 30.2 & 37.2 & 4.7 (162) & 2.8 (12.3)\\
                \textbf{MF-214M} &  \textbf{35.8}  & 55.4 & 38.0 & 21.8 & 38.5 & 46.8 & \textbf{3.9} (162) & 5.7 (15.2) \\
			\hline
	 ResNet18 \cite{he2016deep} &  31.8  & 49.6 & 33.6 & 16.3 & 34.3 & 43.2 &	29 (181) &	11.2 (21.3) \\
	\textbf{MF-294M} &  \textbf{36.6}  & 56.6 & 38.6 & 21.9 & 39.5 & 47.9 & \textbf{5.5} (164) & 6.5 (16.1)\\
		    \hline
    ResNet50 \cite{he2016deep} &  36.5  & 55.4 & 39.1 & 20.4 & 40.3 & 48.1 &	84 (239) &	23.3 (37.7) \\
 PVT-Tiny \cite{wang2021pvtv1} &  36.7  & 56.9 & 38.9 & 22.6 & 38.8 & 50.0 & 70 (221) &  12.3 (23.0) \\			      
 ConT-M \cite{yan2104contnet}  &  37.9  & 58.1 & 40.2 & 23.0 & 40.6 & 50.4 & 65 (217) & 16.8 (27.0) \\			      
   \textbf{MF-508M} &  \textbf{38.0}  & 58.3 & 40.3 & 22.9 & 41.2 & 49.7 & \textbf{9.8} (168) & 8.4 (17.9)\\
			\specialrule{.1em}{.05em}{.05em} 
		\end{tabular}
		}
	\end{center}
	\vspace{-3mm}
	\caption{\textbf{COCO object detection results in RetinaNet framework}. All models are trained on \texttt{train2017} for 12 epochs (1$\times$) from ImageNet pretrained weights, and tested on \texttt{val2017}. We use initial \textbf{MF} (e.g. MF-508M) to refer Mobile-Former. MAdds and \#Params are in the format of ``backbone (total)". MAdds is based on the image size 800$\times$1333.}
	\label{table:coco-det-results}
	\vspace{-1mm}
\end{table}

Object detection experiments are conducted on COCO 2017 \cite{lin2014microsoft}, which contains 118K training and 5K validation images. We evaluate Mobile-Former in two detection frameworks: (a) comparing with other backbone networks in RetinaNet \cite{Lin_2017_ICCV_retinanet_focal} framework that has dense proposals, and (b) end-to-end comparison with DETR \cite{nicolas2020detr} where sparse proposals are used.

\vspace{1mm}
\noindent \textbf{RetinaNet training setup:} We follow the standard settings of RetinaNet \cite{Lin_2017_ICCV_retinanet_focal} and replace the backbone with our Mobile-Former to generate multi-scale feature maps. All models are trained for 12 epochs (1×) from ImageNet pretrained weights.

\vspace{1mm}
\noindent \textbf{DETR training setup:} All Mobile-Former models are trained for 300 epochs on 8 GPUs with 2 images per GPU. AdamW optimizer is used with initial learning rate 1e-5 for the backbone and 1e-4 for the head. The learning rate drops by a factor of 10 after 200 epochs. The weight decay is 1e-4 and dropout rate is 0.1. 
BatchNorm layers in the ImageNet pretrained Mobile-Former backbone are frozen. The head includes 100 object queries with 256 channels. 

\vspace{1mm}
\noindent \textbf{Efficient and effective backbone in RetinaNet:}
In \Tref{table:coco-det-results}, we compare Mobile-Former with both CNNs (ResNet \cite{he2016deep}, MobileNetV3 \cite{Howard_2019_ICCV_mbnetv3}, ShuffleNetV2 \cite{ma_2018_ECCV}) and vision transformers (PVT \cite{wang2021pvtv1} and ConT \cite{yan2104contnet}). Mobile-Former significantly outperforms MobileNetV3 and ShuffleNetV2 by 8.3+ AP under similar computational cost. Compared to ResNet and transformer variants, Mobile-Former achieves higher AP with significantly less FLOPs in the backbone. Specifically, Mobile-Former-508M only takes 9.8G FLOPs in backbone but achieves 38.0 AP, outperforming ResNet-50, PVT-Tiny, and ConT-M which consume 7 times more computation (65G to 84G FLOPs) in the backbone. This showcases Mobile-Former as an effective and efficient backbone in object detection.

\vspace{1mm}
\noindent \textbf{Efficient and effective end-to-end detector:}
\Tref{table:coco-det-detr-cmp} compares end-to-end Mobile-Former detectors (denoted by prefix E2E-MF) with DETR \cite{nicolas2020detr}. All models use 100 object queries. Our E2E-MF-508M gains 1.1 AP over DETR but consumes fewer FLOPs (48\%), fewer model parameters (64\%) and fewer training epochs (300 vs. 500). It is just a little behind DETR-DC5 that has four times more FLOPs. The other three Mobile-Former variants achieve 40.5, 39.3 and 37.2 AP with 24.1G, 17.8G and 12.7G FLOPs respectively, providing more compact end-to-end detectors.

\begin{table}[t!]
	\begin{center}
	    \smallfootnote
	    \setlength{\tabcolsep}{1.0mm}{
		\begin{tabular}{l|ccc|ccc|c|c}
		    \specialrule{.1em}{.05em}{.05em} 
			\multirow{2}{*}{Model} & \multirow{2}{*}{AP} & \multirow{2}{*}{AP\textsubscript{50}} & \multirow{2}{*}{AP\textsubscript{75}} & \multirow{2}{*}{AP\textsubscript{S}} & \multirow{2}{*}{AP\textsubscript{M}} & \multirow{2}{*}{AP\textsubscript{L}} & MAdds & \#Params  \\
	         &  &  &  &  &  &  & (G) & (M)  \\	
			\specialrule{.1em}{.05em}{.05em} 
            DETR \cite{nicolas2020detr} &  42.0  & 62.4 & 44.2 & 20.5 & 45.8 & 61.1 &	86 &	41.3 \\
            DETR-DC5\cite{nicolas2020detr} &  \textbf{43.3}  & \textbf{63.1} & 45.9 & 22.5 & \textbf{47.3} & \textbf{61.1} & 187 &	41.3 \\

		    \hline
            \textbf{E2E-MF-508M} &  43.1  & 61.9 & \textbf{46.8} & \textbf{23.8} & 46.5 & 60.4 & \textbf{41.4} & \textbf{26.6}\\
            \textbf{E2E-MF-294M} &  40.5  & 58.8 & 43.5 & 20.6 & 44.0 & 56.9 & \textbf{24.1} & \textbf{25.1}\\
            \textbf{E2E-MF-214M} &  39.3  & 57.3 & 42.1 & 19.9 & 42.4 & 56.6 & \textbf{17.8} & \textbf{20.1} \\
            \textbf{E2E-MF-151M} &  37.2  & 54.5 & 39.9 & 17.4 & 39.8 & 54.9 & \textbf{12.7} & \textbf{14.8}\\ 
			\specialrule{.1em}{.05em}{.05em} 
		\end{tabular}
		}
	\end{center}
	\vspace{-3mm}
	\caption{\textbf{End-to-end object detection results on COCO}. All models are trained on \texttt{train2017} and tested on \texttt{val2017}. DETR baselines are trained for 500 epochs, while our Mobile-Former models are trained for 300 epochs. We use initial \textbf{E2E-MF} (e.g. E2E-MF-508M) to refer end-to-end Mobile-Former detectors. MAdds is based on image size 800$\times$1333.}
	\label{table:coco-det-detr-cmp}
	\vspace{-2mm}
\end{table}

\vspace{1mm}
\noindent \textbf{Ablations of key components:} \Tref{table:coco-det-detr-ablation} shows the effects of three proposed components in the end-to-end Mobile-Former detector: (a) spatial-aware dynamic ReLU in backbone, (b) multi-scale Mobile-Former head, and (c) adapting position embedding. E2E-MF-508M is used and all models are trained for 300 epochs. Compared to DETR trained in 300 epochs, replacing ResNet-50 backbone with MF-508M saves computations, but results in 1.2 AP drop. Adding spatial-aware dynamic ReLU gains 1.1 AP (39.4 $\rightarrow$ 40.5). Then using multi-scale Mobile-Former head to replace DETR's encoder/decoder gains another 0.9 AP (40.5 $\rightarrow$ 41.4). The detection of small and medium objects is improved, while a slight degradation is found in large objects. 
Finally, adapting position embedding provides additional 1.7 AP gain. The three proposed components are complementary, gaining 3.7 AP (39.4 $\rightarrow$ 43.1) in total.

\begin{table}[t!]
	\begin{center}
	    \smallfootnote
	    \setlength{\tabcolsep}{0.9mm}{
	    \begin{tabular}{cccc|ccc|ccc}
		    \specialrule{.1em}{.05em}{.05em} 
			Backbone & SP-DY & Head & Adapt & AP & AP\textsubscript{50} & AP\textsubscript{75} & AP\textsubscript{S} & AP\textsubscript{M} & AP\textsubscript{L}  \\
			 & ReLU &  & PE & & &  &  &  &  \\
			\specialrule{.1em}{.05em}{.05em}
			ResNet-50 & & DETR & & 40.6 & 61.6 & 42.7 & 19.9 & 44.3 & 60.2  \\
			\hline
			MF-508M & & DETR & & 39.4  & 59.4 & 41.3 & 18.0 & 42.7 & 58.8  \\
        MF-508M & \checkmark & DETR & & 40.5  & 60.7 & 42.4 & 19.0 & 44.0 & 60.0  \\
        MF-508M & \checkmark & MF-Head &  & 41.4 & 60.4 & 43.9 & 21.1 & 45.0 & 59.3  \\
    MF-508M & \checkmark & MF-Head & \checkmark & \textbf{43.1}  & \textbf{61.9} & \textbf{46.8} & \textbf{23.8} & \textbf{46.5} & \textbf{60.4} \\
			\specialrule{.1em}{.05em}{.05em} 
		\end{tabular}
		}
	\end{center}
	\vspace{-3mm}
	\caption{\textbf{Ablation of end-to-end Mobile-Former on COCO object detection}. All models are trained on \texttt{train2017} for 300 epochs and tested on \texttt{val2017}. The first line is the baseline of DETR \cite{nicolas2020detr} trained in 300 epochs. MF-508M refers to Mobile-Former-508M backbone. SP-DY-ReLU refers to spatial-aware dynamic ReLU in backbone. DETR head includes 6 encoder and 6 decoder layers with resolution $\frac{1}{32}$. MF head includes 9 Mobile-Former blocks (5, 2, 2 at resolution $\frac{1}{32}$, $\frac{1}{16}$, and $\frac{1}{8}$ respectively).}
	\label{table:coco-det-detr-ablation}
	\vspace{-2mm}
\end{table}

\section{Limitations and Discussions}
Although Mobile-Former has faster inference than MobileNetV3 \cite{Howard_2019_ICCV_mbnetv3} for large images, it is slower as the image becomes smaller. Please see \Fref{fig:latency} in appendix~\ref{apx:more-results}
for comparison on inference latency between Mobile-Former-214M and MobileNetV3 Large. They have similar FLOPs (214M vs 217M), but Mobile-Former is more accurate. The comparison is performed on multiple image sizes due to the resolution variation across tasks (e.g. classification, detection).  
As image resolution decreases, Mobile-Former loses its leading position to MobileNetV3. 
This is because \textit{Former} and embedding projections in \textit{Mobile$\rightarrow$Former} and \textit{Mobile$\leftarrow$Former} are resolution independent, and their PyTorch implementations are not as efficient as convolution. Thus, the overhead is relative large when image is small, but becomes negligible as image size grows. The runtime performance of Mobile-Former can be further improved by optimizing the implementation of these components. We will investigate these in the future work.

Another limitation is that Mobile-Former is not efficient in parameters especially when performing image classification, due to the parameter-heavy classification head. For instance, the head of Mobile-Former-294M consumes 4.6M of total 11.4M parameters (40\%). This problem is mitigated when switching to object detection, due to the removal of image classification head. In addition, \textit{Former} and two-way bridge are computationally but not parametrically efficient. 


\section{Conclusion}
This paper presents Mobile-Former, a new parallel design of MobileNet and Transformer with two-way bridge in between to communicate. It leverages the efficiency of MobileNet in local processing and the advantage of Transformer in encoding global interaction. This design is not only effective to boost accuracy, but also efficient to save computational cost. It outperforms both efficient CNNs and vision transformer variants with a clear margin on image classification and object detection in the low FLOP regime. Furthermore, we build an end-to-end Mobile-Former detector that outperforms DETR but consumes significantly less computations and parameters. We hope Mobile-Former encourage new design of efficient CNNs and transformers.

{\small
\bibliographystyle{ieee_fullname}
\bibliography{egbib}

\begin{thebibliography}{10}\itemsep=-1pt

\bibitem{nicolas2020detr}
Nicolas Carion, Francisco Massa, Gabriel Synnaeve, Nicolas Usunier, Alexander
  Kirillov, and Sergey Zagoruyko.
\newblock End-to-end object detection with transformers.
\newblock In {\em ECCV}, 2020.

\bibitem{Chen_2020_CVPR_addernet}
Hanting Chen, Yunhe Wang, Chunjing Xu, Boxin Shi, Chao Xu, Qi Tian, and Chang
  Xu.
\newblock Addernet: Do we really need multiplications in deep learning?
\newblock In {\em Proceedings of the IEEE/CVF Conference on Computer Vision and
  Pattern Recognition (CVPR)}, June 2020.

\bibitem{Chen2019DynamicCA}
Yinpeng Chen, Xiyang Dai, Mengchen Liu, Dongdong Chen, Lu Yuan, and Zicheng
  Liu.
\newblock Dynamic convolution: Attention over convolution kernels.
\newblock In {\em IEEE Conference on Computer Vision and Pattern Recognition
  (CVPR)}, 2020.

\bibitem{Chen2020DynamicReLU}
Yinpeng Chen, Xiyang Dai, Mengchen Liu, Dongdong Chen, Lu Yuan, and Zicheng
  Liu.
\newblock Dynamic relu.
\newblock In {\em ECCV}, 2020.

\bibitem{Cubuk_2019_CVPR}
Ekin~D. Cubuk, Barret Zoph, Dandelion Mane, Vijay Vasudevan, and Quoc~V. Le.
\newblock Autoaugment: Learning augmentation strategies from data.
\newblock In {\em Proceedings of the IEEE/CVF Conference on Computer Vision and
  Pattern Recognition (CVPR)}, June 2019.

\bibitem{d2021convit}
St{\'e}phane d'Ascoli, Hugo Touvron, Matthew Leavitt, Ari Morcos, Giulio
  Biroli, and Levent Sagun.
\newblock Convit: Improving vision transformers with soft convolutional
  inductive biases.
\newblock {\em arXiv preprint arXiv:2103.10697}, 2021.

\bibitem{deng2009imagenet}
Jia Deng, Wei Dong, Richard Socher, Li-Jia Li, Kai Li, and Li Fei-Fei.
\newblock Imagenet: A large-scale hierarchical image database.
\newblock In {\em 2009 IEEE conference on computer vision and pattern
  recognition}, pages 248--255. Ieee, 2009.

\bibitem{dong2021cswin}
Xiaoyi Dong, Jianmin Bao, Dongdong Chen, Weiming Zhang, Nenghai Yu, Lu Yuan,
  Dong Chen, and Baining Guo.
\newblock Cswin transformer: A general vision transformer backbone with
  cross-shaped windows.
\newblock {\em arXiv preprint arXiv:2107.00652}, 2021.

\bibitem{dosovitskiy2021vit}
Alexey Dosovitskiy, Lucas Beyer, Alexander Kolesnikov, Dirk Weissenborn,
  Xiaohua Zhai, Thomas Unterthiner, Mostafa Dehghani, Matthias Minderer, Georg
  Heigold, Sylvain Gelly, Jakob Uszkoreit, and Neil Houlsby.
\newblock An image is worth 16x16 words: Transformers for image recognition at
  scale.
\newblock In {\em International Conference on Learning Representations}, 2021.

\bibitem{graham2021levit}
Benjamin Graham, Alaaeldin El-Nouby, Hugo Touvron, Pierre Stock, Armand Joulin,
  Herv\'e J\'egou, and Matthijs Douze.
\newblock Levit: a vision transformer in convnet's clothing for faster
  inference.
\newblock {\em arXiv preprint arXiv:22104.01136}, 2021.

\bibitem{Han_2020_CVPR_ghostnet}
Kai Han, Yunhe Wang, Qi Tian, Jianyuan Guo, Chunjing Xu, and Chang Xu.
\newblock Ghostnet: More features from cheap operations.
\newblock In {\em IEEE/CVF Conference on Computer Vision and Pattern
  Recognition (CVPR)}, June 2020.

\bibitem{NEURIPS2020_e069ea4c}
Kai Han, Yunhe Wang, Qiulin Zhang, Wei Zhang, Chunjing XU, and Tong Zhang.
\newblock Model rubiks cube: Twisting resolution, depth and width for tinynets.
\newblock In H. Larochelle, M. Ranzato, R. Hadsell, M.~F. Balcan, and H. Lin,
  editors, {\em Advances in Neural Information Processing Systems}, volume~33,
  pages 19353--19364. Curran Associates, Inc., 2020.

\bibitem{he2016deep}
Kaiming He, Xiangyu Zhang, Shaoqing Ren, and Jian Sun.
\newblock Deep residual learning for image recognition.
\newblock In {\em Proceedings of the IEEE conference on computer vision and
  pattern recognition}, pages 770--778, 2016.

\bibitem{Hinton2021-part-whole}
Geoffrey~E. Hinton.
\newblock How to represent part-whole hierarchies in a neural network.
\newblock {\em CoRR}, abs/2102.12627, 2021.

\bibitem{Howard_2019_ICCV_mbnetv3}
Andrew Howard, Mark Sandler, Grace Chu, Liang-Chieh Chen, Bo Chen, Mingxing
  Tan, Weijun Wang, Yukun Zhu, Ruoming Pang, Vijay Vasudevan, Quoc~V. Le, and
  Hartwig Adam.
\newblock Searching for mobilenetv3.
\newblock In {\em Proceedings of the IEEE/CVF International Conference on
  Computer Vision (ICCV)}, October 2019.

\bibitem{howard2017mobilenets}
Andrew~G Howard, Menglong Zhu, Bo Chen, Dmitry Kalenichenko, Weijun Wang,
  Tobias Weyand, Marco Andreetto, and Hartwig Adam.
\newblock Mobilenets: Efficient convolutional neural networks for mobile vision
  applications.
\newblock {\em arXiv preprint arXiv:1704.04861}, 2017.

\bibitem{Hu_2018_CVPR}
Jie Hu, Li Shen, and Gang Sun.
\newblock Squeeze-and-excitation networks.
\newblock In {\em The IEEE Conference on Computer Vision and Pattern
  Recognition (CVPR)}, June 2018.

\bibitem{li2021micronet}
Yunsheng Li, Yinpeng Chen, Xiyang Dai, Dongdong Chen, Mengchen Liu, Lu Yuan,
  Zicheng Liu, Lei Zhang, and Nuno Vasconcelos.
\newblock Micronet: Improving image recognition with extremely low flops.
\newblock In {\em International Conference on Computer Vision}, 2021.

\bibitem{Lin_FPN}
T. Lin, P. Dollar, R. Girshick, K. He, B. Hariharan, and S. Belongie.
\newblock Feature pyramid networks for object detection.
\newblock In {\em 2017 IEEE Conference on Computer Vision and Pattern
  Recognition (CVPR)}, pages 936--944, July 2017.

\bibitem{Lin_2017_ICCV_retinanet_focal}
Tsung-Yi Lin, Priya Goyal, Ross Girshick, Kaiming He, and Piotr Dollar.
\newblock Focal loss for dense object detection.
\newblock In {\em Proceedings of the IEEE International Conference on Computer
  Vision (ICCV)}, Oct 2017.

\bibitem{lin2014microsoft}
Tsung-Yi Lin, Michael Maire, Serge Belongie, James Hays, Pietro Perona, Deva
  Ramanan, Piotr Doll{\'a}r, and C~Lawrence Zitnick.
\newblock Microsoft coco: Common objects in context.
\newblock In {\em European conference on computer vision}, pages 740--755.
  Springer, 2014.

\bibitem{liu2021Swin}
Ze Liu, Yutong Lin, Yue Cao, Han Hu, Yixuan Wei, Zheng Zhang, Stephen Lin, and
  Baining Guo.
\newblock Swin transformer: Hierarchical vision transformer using shifted
  windows.
\newblock {\em arXiv preprint arXiv:2103.14030}, 2021.

\bibitem{loshchilov2018decoupled}
Ilya Loshchilov and Frank Hutter.
\newblock Decoupled weight decay regularization.
\newblock In {\em International Conference on Learning Representations}, 2019.

\bibitem{Ma_2020_eccv_WeightNetRT}
Ningning Ma, X. Zhang, J. Huang, and J. Sun.
\newblock Weightnet: Revisiting the design space of weight networks.
\newblock volume abs/2007.11823, 2020.

\bibitem{ma_2018_ECCV}
Ningning Ma, Xiangyu Zhang, Hai-Tao Zheng, and Jian Sun.
\newblock Shufflenet v2: Practical guidelines for efficient cnn architecture
  design.
\newblock In {\em The European Conference on Computer Vision (ECCV)}, September
  2018.

\bibitem{sandler2018mobilenetv2}
Mark Sandler, Andrew Howard, Menglong Zhu, Andrey Zhmoginov, and Liang-Chieh
  Chen.
\newblock Mobilenetv2: Inverted residuals and linear bottlenecks.
\newblock In {\em Proceedings of the IEEE Conference on Computer Vision and
  Pattern Recognition}, pages 4510--4520, 2018.

\bibitem{Srinivas_2021_CVPR_bot}
Aravind Srinivas, Tsung-Yi Lin, Niki Parmar, Jonathon Shlens, Pieter Abbeel,
  and Ashish Vaswani.
\newblock Bottleneck transformers for visual recognition.
\newblock In {\em Proceedings of the IEEE/CVF Conference on Computer Vision and
  Pattern Recognition (CVPR)}, pages 16519--16529, June 2021.

\bibitem{tan-ICML19-efficientnet}
Mingxing Tan and Quoc Le.
\newblock Efficientnet: Rethinking model scaling for convolutional neural
  networks.
\newblock In {\em ICML}, pages 6105--6114, Long Beach, California, USA, 09--15
  Jun 2019.

\bibitem{Tan-bmvc2019-mixconv}
Mingxing Tan and Quoc~V. Le.
\newblock Mixconv: Mixed depthwise convolutional kernels.
\newblock In {\em 30th British Machine Vision Conference 2019}, 2019.

\bibitem{Tan_2020_CVPR}
Mingxing Tan, Ruoming Pang, and Quoc~V. Le.
\newblock Efficientdet: Scalable and efficient object detection.
\newblock In {\em Proceedings of the IEEE/CVF Conference on Computer Vision and
  Pattern Recognition (CVPR)}, June 2020.

\bibitem{tolstikhin2021mlpmixer}
Ilya Tolstikhin, Neil Houlsby, Alexander Kolesnikov, Lucas Beyer, Xiaohua Zhai,
  Thomas Unterthiner, Jessica Yung, Andreas Steiner, Daniel Keysers, Jakob
  Uszkoreit, Mario Lucic, and Alexey Dosovitskiy.
\newblock Mlp-mixer: An all-mlp architecture for vision, 2021.

\bibitem{touvron2020deit}
Hugo Touvron, Matthieu Cord, Matthijs Douze, Francisco Massa, Alexandre
  Sablayrolles, and Herv\'e J\'egou.
\newblock Training data-efficient image transformers and distillation through
  attention.
\newblock {\em arXiv preprint arXiv:2012.12877}, 2020.

\bibitem{touvron2021going}
Hugo Touvron, Matthieu Cord, Alexandre Sablayrolles, Gabriel Synnaeve, and
  Hervé Jégou.
\newblock Going deeper with image transformers, 2021.

\bibitem{vahid_2020_CVPR}
Keivan~Alizadeh vahid, Anish Prabhu, Ali Farhadi, and Mohammad Rastegari.
\newblock Butterfly transform: An efficient fft based neural architecture
  design.
\newblock In {\em Proceedings of the IEEE/CVF Conference on Computer Vision and
  Pattern Recognition (CVPR)}, June 2020.

\bibitem{Vaswani_2021_CVPR_halo}
Ashish Vaswani, Prajit Ramachandran, Aravind Srinivas, Niki Parmar, Blake
  Hechtman, and Jonathon Shlens.
\newblock Scaling local self-attention for parameter efficient visual
  backbones.
\newblock In {\em Proceedings of the IEEE/CVF Conference on Computer Vision and
  Pattern Recognition (CVPR)}, pages 12894--12904, June 2021.

\bibitem{NIPS2017_transformer}
Ashish Vaswani, Noam Shazeer, Niki Parmar, Jakob Uszkoreit, Llion Jones,
  Aidan~N Gomez, \L~ukasz Kaiser, and Illia Polosukhin.
\newblock Attention is all you need.
\newblock In I. Guyon, U.~V. Luxburg, S. Bengio, H. Wallach, R. Fergus, S.
  Vishwanathan, and R. Garnett, editors, {\em Advances in Neural Information
  Processing Systems}, volume~30. Curran Associates, Inc., 2017.

\bibitem{wang2021pvtv1}
Wenhai Wang, Enze Xie, Xiang Li, Deng-Ping Fan, Kaitao Song, Ding Liang, Tong
  Lu, Ping Luo, and Ling Shao.
\newblock Pyramid vision transformer: A versatile backbone for dense prediction
  without convolutions, 2021.

\bibitem{rw2019timm}
Ross Wightman.
\newblock Pytorch image models.
\newblock \url{https://github.com/rwightman/pytorch-image-models}, 2019.

\bibitem{wu2021cvt}
Haiping Wu, Bin Xiao, Noel Codella, Mengchen Liu, Xiyang Dai, Lu Yuan, and Lei
  Zhang.
\newblock Cvt: Introducing convolutions to vision transformers, 2021.

\bibitem{Xiao-2021-early-cnns-help-transformers}
Tete Xiao, Mannat Singh, Eric Mintun, Trevor Darrell, Piotr Doll{\'{a}}r, and
  Ross~B. Girshick.
\newblock Early convolutions help transformers see better.
\newblock {\em CoRR}, abs/2106.14881, 2021.

\bibitem{xu2021coscale}
Weijian Xu, Yifan Xu, Tyler Chang, and Zhuowen Tu.
\newblock Co-scale conv-attentional image transformers, 2021.

\bibitem{yan2104contnet}
Haotian Yan, Zhe Li, Weijian Li, Changhu Wang, Ming Wu, and Chuang Zhang.
\newblock Contnet: Why not use convolution and transformer at the same time?
\newblock {\em CoRR}, abs/2104.13497, 2021.

\bibitem{Yang2019CondConvCP}
Brandon Yang, Gabriel Bender, Quoc~V. Le, and Jiquan Ngiam.
\newblock Condconv: Conditionally parameterized convolutions for efficient
  inference.
\newblock In {\em NeurIPS}, 2019.

\bibitem{yuan2021tokens}
Li Yuan, Yunpeng Chen, Tao Wang, Weihao Yu, Yujun Shi, Francis~EH Tay, Jiashi
  Feng, and Shuicheng Yan.
\newblock Tokens-to-token vit: Training vision transformers from scratch on
  imagenet.
\newblock {\em arXiv preprint arXiv:2101.11986}, 2021.

\bibitem{zhang2018mixup}
Hongyi Zhang, Moustapha Cisse, Yann~N. Dauphin, and David Lopez-Paz.
\newblock mixup: Beyond empirical risk minimization.
\newblock In {\em International Conference on Learning Representations}, 2018.

\bibitem{Zhang_2018_CVPR}
Xiangyu Zhang, Xinyu Zhou, Mengxiao Lin, and Jian Sun.
\newblock Shufflenet: An extremely efficient convolutional neural network for
  mobile devices.
\newblock In {\em The IEEE Conference on Computer Vision and Pattern
  Recognition (CVPR)}, June 2018.

\bibitem{zhong2020random}
Zhun Zhong, Liang Zheng, Guoliang Kang, Shaozi Li, and Yi Yang.
\newblock Random erasing data augmentation.
\newblock In {\em Proceedings of the AAAI Conference on Artificial Intelligence
  (AAAI)}, 2020.

\bibitem{Daquan_2020_ECCV_RethinkingBS}
Daquan Zhou, Qi-Bin Hou, Y. Chen, Jiashi Feng, and S. Yan.
\newblock Rethinking bottleneck structure for efficient mobile network design.
\newblock In {\em ECCV}, August 2020.

\bibitem{zhou2021deepvit}
Daquan Zhou, Bingyi Kang, Xiaojie Jin, Linjie Yang, Xiaochen Lian, Zihang
  Jiang, Qibin Hou, and Jiashi Feng.
\newblock Deepvit: Towards deeper vision transformer, 2021.

\bibitem{zhou2021refiner}
Daquan Zhou, Yujun Shi, Bingyi Kang, Weihao Yu, Zihang Jiang, Yuan Li, Xiaojie
  Jin, Qibin Hou, and Jiashi Feng.
\newblock Refiner: Refining self-attention for vision transformers, 2021.

\end{thebibliography}
}

\clearpage
\appendix


\section{Mobile-Former Architecture} \label{apx:arch-all}
\begin{table*}[b!]
	\begin{center}
	    \footnotesize
	    \setlength{\tabcolsep}{0.7mm}{
		\begin{tabular}{c|ccc|ccc|ccc|ccc|ccc|ccc}
		    \specialrule{.1em}{.05em}{.05em} 
			\multirow{2}{*}{Stage}& \multicolumn{3}{c|}{\textbf{Mobile-Former-508M}} & \multicolumn{3}{c|}{\textbf{Mobile-Former-294M}} & \multicolumn{3}{c|}{\textbf{Mobile-Former-214M}} & \multicolumn{3}{c|}{\textbf{Mobile-Former-151M}} & \multicolumn{3}{c|}{\textbf{Mobile-Former-96M}} & \multicolumn{3}{c}{\textbf{Mobile-Former-52M}}  \\
			\cline{2-19}
			 & Block & \#exp & \#out & Block & \#exp & \#out & Block & \#exp & \#out & Block & \#exp & \#out & Block & \#exp & \#out & Block & \#exp & \#out \\
		
			\specialrule{.1em}{.05em}{.05em} 
			token & \multicolumn{3}{c|}{6$\times$192} & \multicolumn{3}{c|}{6$\times$192} & \multicolumn{3}{c|}{6$\times$192} & \multicolumn{3}{c|}{6$\times$192} & \multicolumn{3}{c|}{4$\times$128} & \multicolumn{3}{c}{3$\times$128}  \\
			\hline
			stem & conv 3$\times$3 & -- & 24 &  conv 3$\times$3 & -- & 16 &  conv 3$\times$3 & -- & 12 &  conv 3$\times$3 & -- & 12 &  conv 3$\times$3 & -- & 12 &  conv 3$\times$3 & -- & 8 \\
			
			\hline
			1 &  bneck-lite & 48 & 24 &  bneck-lite & 32 & 16 &  bneck-lite & 24 & 12 &  bneck-lite & 24 & 12 &  bneck-lite & 24 & 12 &   &  &  \\
			\hline
			\multirow{2}{*}{2} &  M-F$^{\downarrow}$ & 144 & 40 &  M-F$^{\downarrow}$ & 96 & 24 &  M-F$^{\downarrow}$ & 72 & 20 &  M-F$^{\downarrow}$ & 72 & 16 &  M-F$^{\downarrow}$ & 72 & 16 &  bneck-lite$^{\downarrow}$ & 24 & 12 \\
			&  M-F & 120 & 40 &  M-F & 96 & 24 &  M-F & 60 & 20 &  M-F & 48 & 16 &   &  &  &  M-F & 36 & 12 \\
			\hline
			\multirow{2}{*}{3} &  M-F$^{\downarrow}$ & 240 & 72 &  M-F$^{\downarrow}$ & 144 & 48 &  M-F$^{\downarrow}$ & 120 & 40 &  M-F$^{\downarrow}$ & 96 & 32 &  M-F$^{\downarrow}$ & 96 & 32 &  M-F$^{\downarrow}$ & 72 & 24 \\
			&  M-F  & 216 & 72 &  M-F  & 192 & 48 &  M-F & 160 & 40 &  M-F & 96 & 32 &  M-F & 96 & 32 &  M-F & 72 & 24 \\
			\hline
			\multirow{4}{*}{4} &  M-F$^{\downarrow}$ & 432 & 128 &  M-F$^{\downarrow}$ & 288 & 96 &  M-F$^{\downarrow}$ & 240 & 80 &  M-F$^{\downarrow}$ & 192 & 64 &  M-F$^{\downarrow}$ & 192 & 64 &  M-F$^{\downarrow}$ & 144 & 48 \\
			
			&  M-F & 512 & 128 &  M-F & 384 & 96 &  M-F & 320 & 80 &  M-F & 256 & 64 &  M-F & 256 & 64 &  M-F & 192 & 48 \\
			
			&  M-F & 768 & 176 &  M-F & 576 & 128 &  M-F & 480 & 112 &  M-F & 384 & 88 &  M-F & 384 & 88 &  M-F & 288 & 64 \\
			
			&  M-F & 1056 & 176 &  M-F & 768 & 128 &  M-F & 672 & 112 &  M-F & 528 & 88 &  &  &  &   &  &  \\

			\hline
			\multirow{4}{*}{5} &  M-F$^{\downarrow}$ & 1056 & 240 &  M-F$^{\downarrow}$ & 768 & 192 &  M-F$^{\downarrow}$ & 672 & 160 &  M-F$^{\downarrow}$ & 528 & 128 &  M-F$^{\downarrow}$ & 528 & 128 &  M-F$^{\downarrow}$ & 384 & 96 \\
			
			&  M-F & 1440 & 240 &  M-F & 1152 & 192 &  M-F & 960 & 160 &  M-F & 768 & 128 &  M-F & 768 & 128 &  M-F & 576 & 96 \\
			
			&  M-F & 1440 & 240 &  M-F & 1152 & 192 &  M-F & 960 & 160 &  M-F & 768 & 128 &  conv 1$\times$1 & -- & 768 &  conv 1$\times$1 & -- & 576  \\
			
			& conv 1$\times$1 & -- & 1440 &  conv 1$\times$1 & -- & 1152 &  conv 1$\times$1 & -- & 960 &  conv 1$\times$1 & -- & 768 &   &  &  &   &  & \\
			
			\specialrule{.1em}{.05em}{.05em} 
			pool & \multirow{2}{*}{--} & \multirow{2}{*}{--} & \multirow{2}{*}{1632} &  \multirow{2}{*}{--} & \multirow{2}{*}{--} & \multirow{2}{*}{1344} &  \multirow{2}{*}{--} & \multirow{2}{*}{--} & \multirow{2}{*}{1152}  &  \multirow{2}{*}{--} & \multirow{2}{*}{--} & \multirow{2}{*}{960} & \multirow{2}{*}{--}  &\multirow{2}{*}{--}  & \multirow{2}{*}{896} & \multirow{2}{*}{--}  & \multirow{2}{*}{--} & \multirow{2}{*}{704}\\
			concat &  &  &  &   &  &  &   &  &  & & &  &   &  & &   &  & \\
			\hline
			FC1 & -- & -- & 1920 &  -- & -- & 1920 &  -- & -- & 1600 &  -- & -- & 1280 & --  & -- & 1280 & --  & -- & 1024 \\
			FC2 & -- & -- & 1000 &  -- & -- & 1000 &  -- & -- & 1000 &  -- & -- & 1000 & --  & -- & 1000 & --  & -- & 1000 \\
			\specialrule{.1em}{.05em}{.05em} 
		\end{tabular}
		}
	\end{center}
	\vspace{-1mm}
	\caption{\textbf{Specification of Mobile-Former models}. ``bneck-lite" denotes the lite bottleneck block \cite{li2021micronet}. ``bneck-lite$^{\downarrow}$" denotes the downsample variant of lite bottleneck, in which the depthwise convolution has stride 2. ``M-F" denotes the Mobile-Former block and ``M-F$^{\downarrow}$" denotes the Mobile-Former block for downsampling. Mobile-Former-26M has a similar architecture to Mobile-Former-52M except replacing all 1$\times$1 convolutions with group convolution (group=4).}
	\vspace{-1mm}
	\label{table:mf-achs}
\end{table*}

\noindent \textbf{Seven model variants:} 
\Tref{table:mf-achs} shows six Mobile-Former models (508M--52M). The smallest model Mobile-Former-26M has similar architecture to Mobile-Former-52M except replacing all 1$\times$1 convolutions with group convolution (group=4). They are used either in image classification or as the backbone of object detectors. These models are manually designed without searching for the optimal architecture parameters (e.g. width or depth). We follow the well known rules used in MobileNet~\cite{sandler2018mobilenetv2, Howard_2019_ICCV_mbnetv3} : (a) the number of channels increases across stages, and (b) the channel expansion rate starts from three at low levels and increases to six at high levels. For the four bigger models (508M--151M), we use six global tokens with dimension 192 and eleven Mobile-Former blocks. But these four models have different widths. Mobile-Former-96M and Mobile-Former-52M are shallower (with only eight Mobile-Former blocks) to meet the low computational budget. 

\vspace{2mm}
\noindent \textbf{Downsample Mobile-Former block:} 
Note that stage 2--5 has a downsample variant of Mobile-Former block (denoted as M-F$^{\downarrow}$ in \Tref{table:mf-achs}) to handle the spatial downsampling. 
M-F$^{\downarrow}$ has a slightly different \textit{Mobile} sub-block that includes four (instead of three) convolutional layers (depthwise$\rightarrow$pointwise$\rightarrow$depthwise$\rightarrow$pointwise), where the first depthwise convolution layer has stride two. The number of channels expands in each depthwise convolution, and squeezes in the following pointwise convolution. This saves computations as the two costly pointwise convolutions are performed at the lower resolution after downsampling.

\vspace{2mm}
\noindent \textbf{Training hyper-parameters:}
\begin{table}
	\begin{center}
	    \footnotesize
	    \setlength{\tabcolsep}{2.3mm}{
		\begin{tabular}{l|ccc }
		    \specialrule{.1em}{.05em}{.05em} 
			Model & Learing Rate & Weight Decay & Dropout \\
		
			\specialrule{.1em}{.05em}{.05em} 
			
			Mobile-Former-26M  & 8e-4 & 0.08 & 0.1  \\
			
			Mobile-Former-52M & 8e-4 & 0.10 & 0.2  \\
			
			Mobile-Former-96M & 8e-4 & 0.10 & 0.2  \\
			
			Mobile-Former-151M & 9e-4 & 0.10 & 0.2  \\
			
			Mobile-Former-214M  & 9e-4 & 0.15 & 0.2 \\
			
			Mobile-Former-294M & 1e-3 & 0.20 & 0.3  \\
			
			Mobile-Former-508M & 1e-3 & 0.20 & 0.3 \\ 	
			\specialrule{.1em}{.05em}{.05em} 
		\end{tabular}
		}
	\end{center}
	\vspace{-2mm}
	\caption{\textbf{Hyper-parameters} of seven Mobile-Former models for ImageNet \cite{deng2009imagenet} classification. }
	\label{table:imagenet-hyper}
\end{table}
\Tref{table:imagenet-hyper} lists three hyper-parameters (initial learning rate, weight decay and dropout rate) used for training Mobile-Former models in ImageNet classification. Their values increase as the model becomes bigger to prevent overfitting. Our implementation is based on timm framework \cite{rw2019timm}.

\begin{table}
	\begin{center}
	    \footnotesize
	    \setlength{\tabcolsep}{0.85mm}{
		\begin{tabular}{c|cc|cc|cc|cc}
		    \specialrule{.1em}{.05em}{.05em} 
			\multirow{2}{*}{Stage}& \multicolumn{2}{c|}{\textbf{E2E-MF}} & \multicolumn{2}{c|}{\textbf{E2E-MF}} & \multicolumn{2}{c|}{\textbf{E2E-MF}} & \multicolumn{2}{c}{\textbf{E2E-MF}}   \\
			& \multicolumn{2}{c|}{\textbf{508M}} & \multicolumn{2}{c|}{\textbf{294M}} & \multicolumn{2}{c|}{\textbf{214M}} & \multicolumn{2}{c}{\textbf{151M}}   \\
		
			\specialrule{.1em}{.05em}{.05em} 
			query & \multicolumn{2}{c|}{100$\times$256} & \multicolumn{2}{c|}{100$\times$256} & \multicolumn{2}{c|}{100$\times$256} & \multicolumn{2}{c}{100$\times$256}   \\
			\hline
			\multirow{2}{*}{$\frac{1}{32}$} & projection & & projection & & projection & & projection & \\
			& $^{\dag}$M-F & $\times$5 & $^{\dag}$M-F & $\times$6  & $^{\dag}$M-F & $\times$5 &  $^{\dag}$M-F & $\times$3 \\
		
			\hline
			\multirow{2}{*}{$\frac{1}{16}$} & up-conv & & up-conv && up-conv && up-conv &\\
			 & M-F & $\times$2 & M-F & $\times$3  & M-F &$\times$2 &  M-F& $\times$2 \\
		    \hline
		    \multirow{2}{*}{$\frac{1}{8}$} & up-conv & &\multirow{2}{*}{--}  & &\multirow{2}{*}{--} & & \multirow{2}{*}{--} &\\	
			 & M-F &$\times$2 & & & &&  \\	
			\specialrule{.1em}{.05em}{.05em} 
		\end{tabular}
		}
	\end{center}
	\vspace{-1mm}
	\caption{\textbf{Specification of head variants} in end-to-end Mobile-Former object detectors. 100 object queries with dimension 256 are used. ``projection" denotes projecting an input feature map linearly to 256 channels (through a 1$\times$1 convolution). ``up-conv" denotes a convolutional block for upsampling that includes bilinear interpolation followed by a 3$\times$3 depthwise and a pointwise convolution. ``M-F $\times$2" refers to stacking two Mobile-Former blocks. In the detection head, we use lite bottleneck \cite{li2021micronet} in \textit{Mobile} sub-block to reduce the computational cost. At the lowest resolution $\frac{1}{32}$, multi-head attention is added into \textit{Mobile}, which is denoted as $^{\dag}$M-F. } 
	\vspace{-1mm}
	\label{table:e2e-head-achs}
\end{table}

\vspace{2mm}
\noindent \textbf{Head model variants in end-to-end object detection:}
\Tref{table:e2e-head-achs} shows the head structures for four end-to-end Mobile-Former detectors. All share similar structure and have 100 object queries with dimension 256. The largest model (E2E-MF-508M) has the heaviest head with 9 Mobile-Former blocks over three scales, while the other three smaller models have 9, 7, 5 blocks respectively over two scales to save computations. 

All models start by projecting the input feature map linearly to 256 channels through a 1$\times$1 convolution. Then multiple Mobile-Former blocks are stacked with upsampling block in between to move upscale. The upsampling block (denoted as ``up-conv") includes three steps: (a) increasing feature resolution by two using bilinear interpolation, (b) adding the feature output from the backbone, and (c) applying a 3$\times$3 depthwise and a pointwise convolution. To handle the computational boost due to resolution increasing,  we use lite bottleneck \cite{li2021micronet} in \textit{Mobile}. Moreover, we find that the performance can be further improved at a small additional cost by adding multi-head attention in \textit{Mobile} sub-block at the lowest scale ($\frac{1}{32}$) of the head (denoted as $^{\dag}$M-F). It is especially helpful for detecting large objects. 

\section{More Experimental Results} \label{apx:more-results}

\noindent \textbf{Inference latency:}
\begin{figure}[t]
	\begin{center}
		\includegraphics[width=1.0\linewidth]{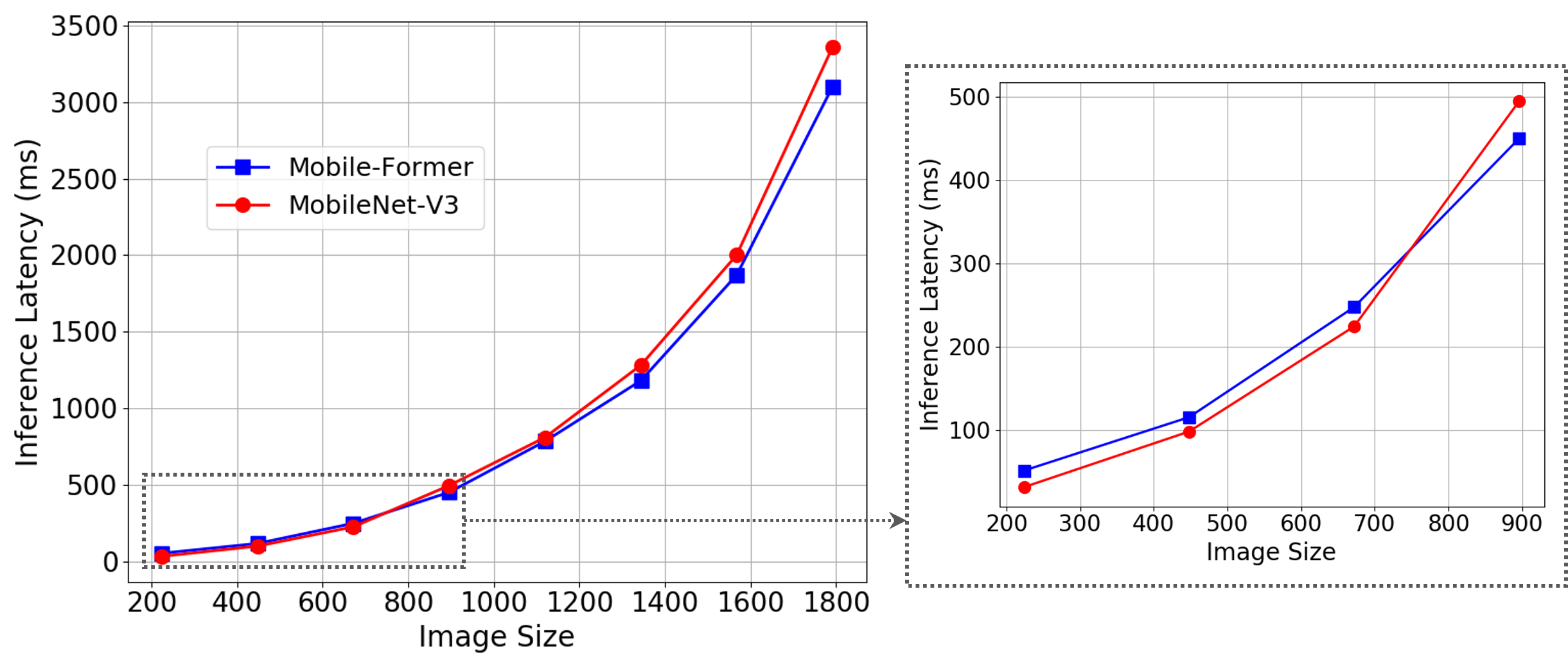}
	\end{center}
	\vspace{-3mm}
	\caption{\textbf{Inference latency} over different image sizes. The latency is measured on an Intel(R) Xeon(R) CPU E5-2650 v3 (2.3GHz), following the common settings (single-thread with batch size 1) in \cite{sandler2018mobilenetv2, Howard_2019_ICCV_mbnetv3}. Mobile-Former-214M is compared with MobileNetV3 Large \cite{Howard_2019_ICCV_mbnetv3} as they have similar FLOPs (214M vs. 217M). Mobile-Former is slower when image size is small, but has faster inference than MobileNetV3 as image size grows above 750$\times$750. Best viewed in color.}
	\label{fig:latency}
	\vspace{-2mm}
\end{figure}
\Fref{fig:latency} compares between Mobile-Former-214M and MobileNetV3 Large \cite{Howard_2019_ICCV_mbnetv3} on inference latency, as they have similar FLOPs (214M vs. 217M). The latency is measured on an Intel(R) Xeon(R) CPU E5-2650 v3 (2.3GHz), following the common settings (single-thread with batch size 1) in \cite{sandler2018mobilenetv2, Howard_2019_ICCV_mbnetv3}. 
The comparison is performed on multiple image sizes due to the resolution variation across tasks (e.g. classification, detection). Mobile-Former is behind MobileNetV3 at low resolution (224$\times$224). As the image resolution increases, the gap shrinks until resolution 750$\times$750, after which Mobile-Former has faster inference. 

This is because \textit{Former} and embedding projections in \textit{Mobile$\rightarrow$Former} and \textit{Mobile$\leftarrow$Former} are resolution independent, and their PyTorch implementations are not as efficient as convolution. Thus, the overhead is relative large when image is small, but becomes negligible as image size grows. The runtime performance of Mobile-Former can be further improved by optimizing the implementation of these components. We will investigate this in the future work.

\begin{table}[t!]
	\begin{center}
	    \footnotesize
	    \setlength{\tabcolsep}{2.8mm}{
		\begin{tabular}{c|cc|cc}
		    \specialrule{.1em}{.05em}{.05em} 
			Kernel Size in Mobile & \#Param & MAdds &Top-1 &	Top-5   \\
		
			\specialrule{.1em}{.05em}{.05em} 
			3$\times$3  & 11.4M & 294M & 77.8 &  93.7		 \\
			5$\times$5 & 11.5M & 332M & 77.9 & 93.9		 \\
			
			\specialrule{.1em}{.05em}{.05em} 
		\end{tabular}
		}
	\end{center}
	\vspace{-2mm}
	\caption{\textbf{Ablation of the kernel size in the depthwise convolution} (in \textit{Mobile} sub-block). The evaluation is performed on ImageNet \cite{deng2009imagenet} classification. Mobile-Former-294M is used.}
	\label{table:ablation-kernel-size}
\end{table}

\begin{table}[t!]
	\begin{center}
	    \smallfootnote
	    \setlength{\tabcolsep}{1.0mm}{
		\begin{tabular}{c|ccc|ccc|c|c}
		    \specialrule{.1em}{.05em}{.05em} 
			MHA in \textit{Mobile}  & \multirow{2}{*}{AP} & \multirow{2}{*}{AP\textsubscript{50}} & \multirow{2}{*}{AP\textsubscript{75}} & \multirow{2}{*}{AP\textsubscript{S}} & \multirow{2}{*}{AP\textsubscript{M}} & \multirow{2}{*}{AP\textsubscript{L}} & MAdds & \#Params  \\
	         at scale $\frac{1}{32}$ &  &  &  &  &  &  & (G) & (M)  \\	
			\specialrule{.1em}{.05em}{.05em} 
              &  42.5  & 61.0 & 46.0 & 23.2 & 46.3 & 58.7 & \textbf{36.0} &	\textbf{23.7} \\

		    \hline
            \checkmark &  \textbf{43.1}  & \textbf{61.9} & \textbf{46.8} & \textbf{23.8} & \textbf{46.5} & \textbf{60.4} & 41.4 & 26.6\\
            
			\specialrule{.1em}{.05em}{.05em} 
		\end{tabular}
		}
	\end{center}
	\vspace{-3mm}
	\caption{\textbf{Ablation of multi-head attention (MHA)} in \textit{Mobile} at resolution $\frac{1}{32}$ of the detection head. The evaluation is performed on COCO \cite{lin2014microsoft} object detection. Both models are trained on \texttt{train2017} for 300 epochs and tested on \texttt{val2017}. E2E-MF-508M is used. MAdds is based on image size 800$\times$1333.}
	\label{table:coco-det-detr-mha-mf}
	\vspace{-2mm}
\end{table}

\vspace{2mm}
\noindent \textbf{Ablation of the kernel size in \textit{Mobile}:}
We perform an ablation on the kernel size of the depthwise convolution in \textit{Mobile}, to validate the contribution of \textit{Former} and bridge on global interaction. \Tref{table:ablation-kernel-size} shows that the gain of increasing kernel size (from 3$\times$3 to 5$\times$5) is negligible. We believe this is because \textit{Former} and the bridge enlarge the reception field for \textit{Mobile} via fusing global features. Therefore, using larger kernel size is not necessary in Mobile-Former.

\vspace{2mm}
\noindent \textbf{Ablation of multi-head attention in \textit{Mobile} at resolution $\frac{1}{32}$ of the detection head:} \Tref{table:coco-det-detr-mha-mf} shows the effect of using multi-head attention (MHA) in the five blocks at the lowest resolution $\frac{1}{32}$ ($^{\dag}$M-F in \Tref{table:e2e-head-achs}). Without MHA, a solid performance (42.5 AP) is achieved at low FLOPs (36.0G). Adding MHA gains 0.6 AP with 15\% additional computational cost. It is especially helpful for detecting large objects (58.7$\rightarrow$60.4 AP\textsubscript{L}). 

\section{Visualization} \label{apx:vis}

\begin{figure}
	\begin{center}
		\includegraphics[width=1.0\linewidth]{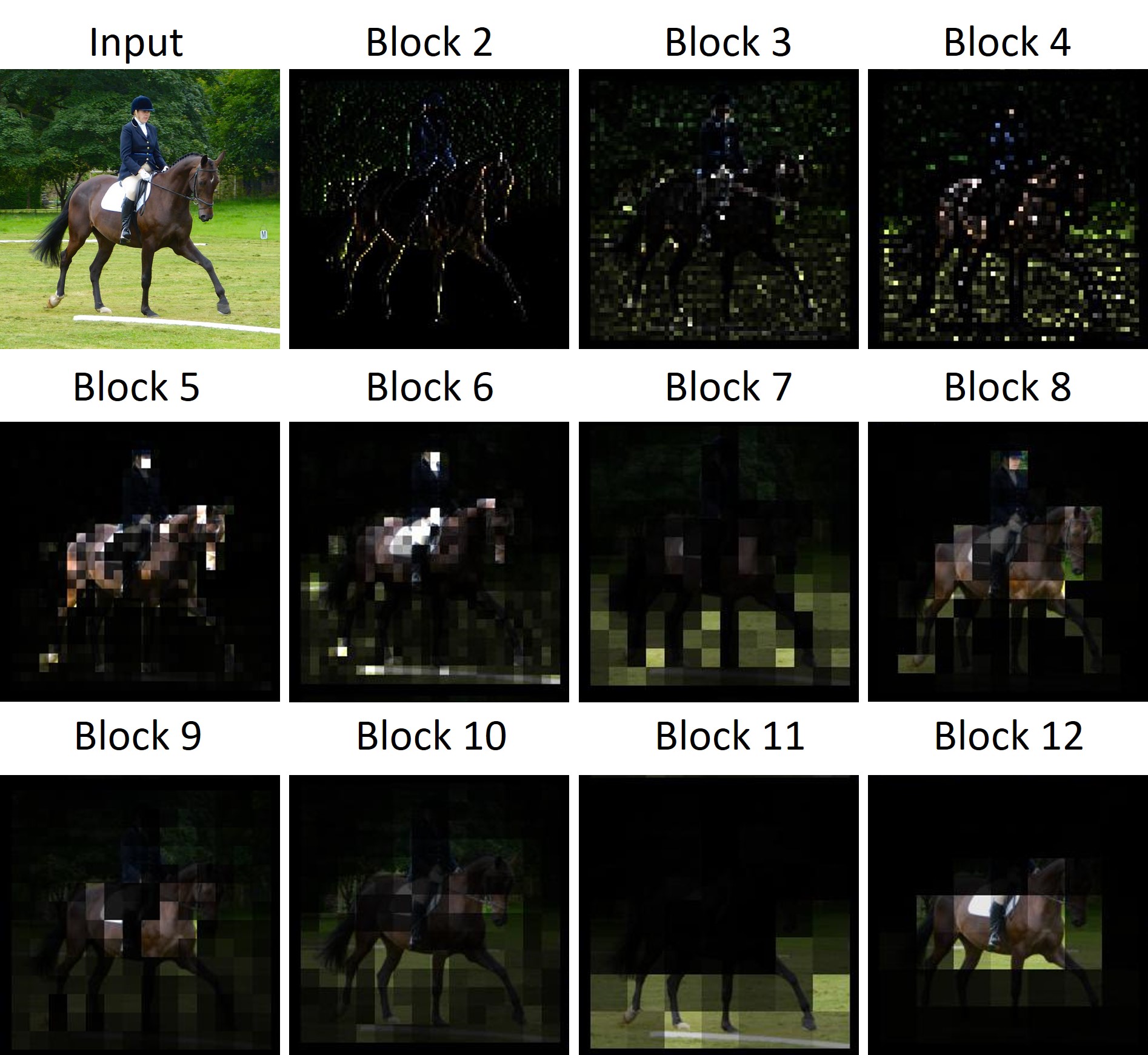}
	\end{center}
	\vspace{-4mm}
	\caption{\textbf{Cross attention over the entire featuremap for the first token in \textit{Mobile$\rightarrow$Former}} across all Mobile-Former blocks. Attention is normalized over pixels, showing the focused region. The focused region changes from low to high levels. The token starts paying more attention to edges/corners at block 2--4. Then it focuses more on a large region rather than scattered small pieces at block 5--12. The focused region shifts between the foreground (person and horse) and background (grass). Finally, it locks the most discriminative part (horse body and head) for classification. Best viewed in color.}
	\label{fig:vis-m2f}
\end{figure}

\begin{figure}
	\begin{center}
		\includegraphics[width=1.0\linewidth]{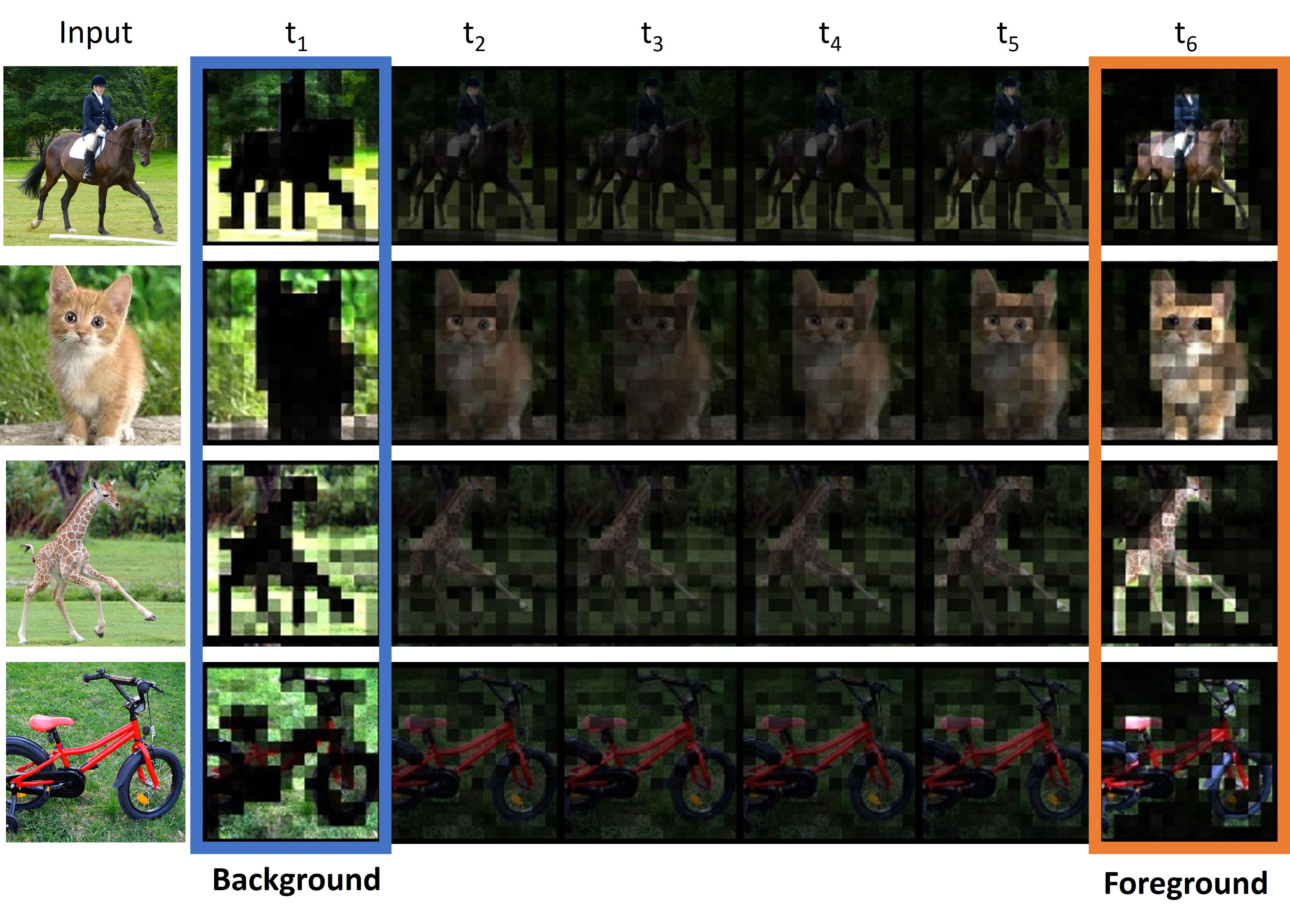}
	\end{center}
	\vspace{-4mm}
	\caption{\textbf{Cross attention in \textit{Mobile$\leftarrow$Former} separates foreground and background at middle layers}. Attention is normalized over tokens showing the contribution of different tokens at each pixel. Block 8 is chosen where background pixels pay more attention to the first token and foreground pixels pay more attention to the last token. Best viewed in color.
	}
	\label{fig:vis-f2m}
\end{figure}

\begin{figure*}
	\begin{center}
		\includegraphics[width=1.0\linewidth]{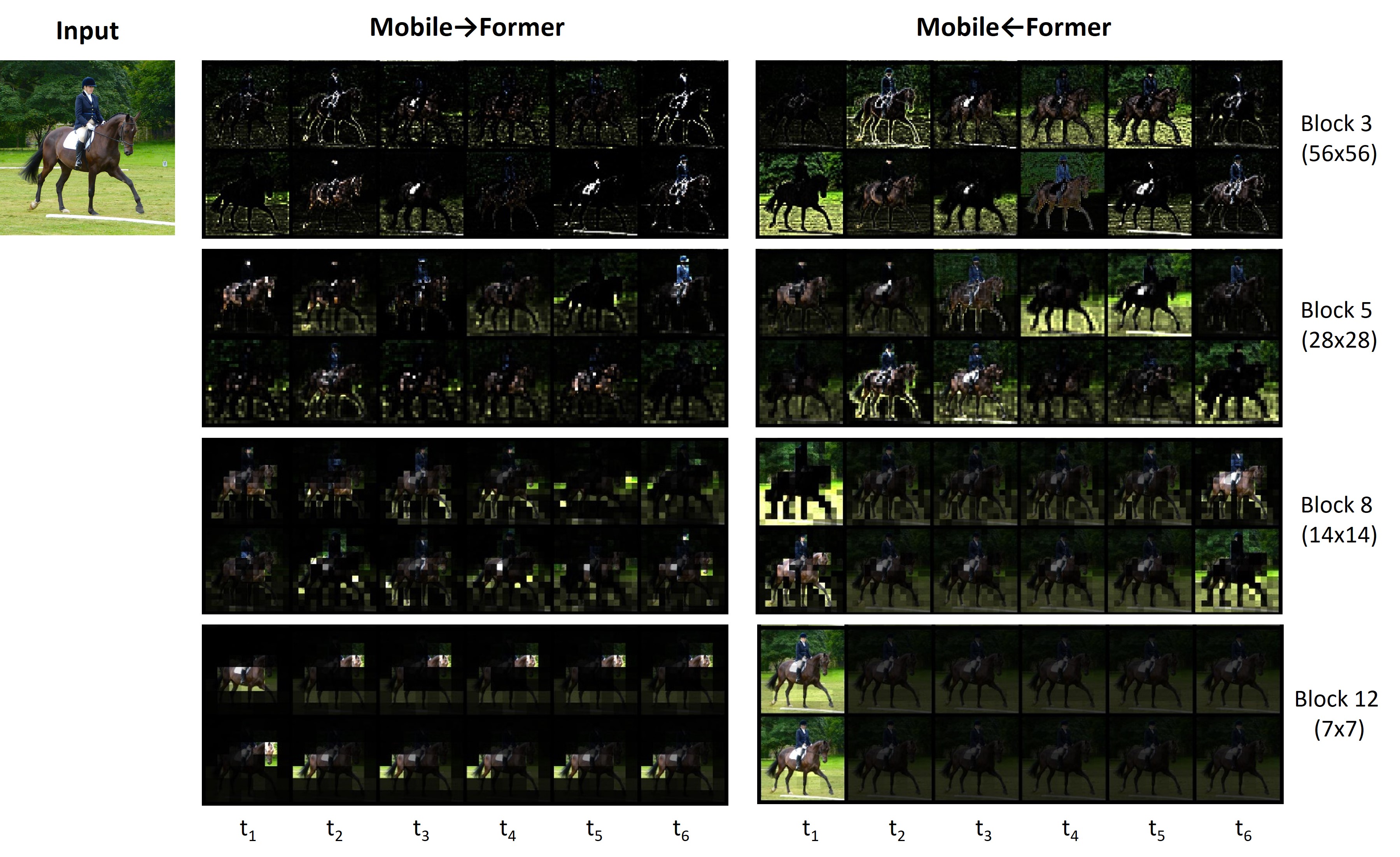}
	\end{center}
	\vspace{-4mm}
	\caption{\textbf{Visualization of the two-way cross attention}: \textit{Mobile$\rightarrow$Former} and \textit{Mobile$\leftarrow$Former}. Mobile-Former-294M is used, which includes six tokens (each corresponds to a column). Four blocks with different input resolutions are selected and each has two attention heads that are visualized in two rows. Attention in \textit{Mobile$\rightarrow$Former} (left half) is normalized over pixels, showing the focused region per token. Attention in \textit{Mobile$\leftarrow$Former} (right half) is normalized over tokens showing the contribution of different tokens at each pixel. The cross attention has less variation across tokens at high levels than low levels. Specifically, token 2--5 in the last block have very similar cross attention. Best viewed in color.}
	\label{fig:vis}
\end{figure*}

In order to understand the collaboration between \textit{Mobile} and \textit{Former}, we visualize the cross attention on the two-way bridge (i.e. \textit{Mobile$\rightarrow$Former} and \textit{Mobile$\leftarrow$Former}) in \Fref{fig:vis-m2f}, \ref{fig:vis-f2m}, and \ref{fig:vis}. The ImageNet pretrained Mobile-Former-294M is used, which includes six global tokens and eleven Mobile-Former blocks. We observe three interesting patterns as follows: 

\vspace{2mm}
\noindent \textbf{Patten 1 -- global tokens shift focus over levels:}
 The focused regions of global tokens change progressively from low to high levels. \Fref{fig:vis-m2f} shows the cross attention over pixels for the first token in \textit{Mobile$\rightarrow$Former}. This token begins focusing on local features, e.g. edges/corners (at block 2-4). Then it pays more attention to regions with connected pixels. Interestingly, the focused region shifts between foreground (person and horse) and background (grass) across blocks. Finally, it locates the most discriminative region (horse body and head) for classification.

 \vspace{2mm}
\noindent \textbf{Pattern 2 -- foreground and background are separated in middle layers:}
 The separation between foreground and background is surprisingly found in \textit{Mobile$\leftarrow$Former} at middle layers (e.g. block 8). \Fref{fig:vis-f2m} shows the cross attention over six tokens for each pixel in the featuremap. Clearly, the foreground and background are separated in the first and last tokens. This shows that global tokens learn meaningful prototypes that cluster pixels with similar semantics.
 
\begin{figure*}
	\begin{center}
		\includegraphics[width=0.96\linewidth]{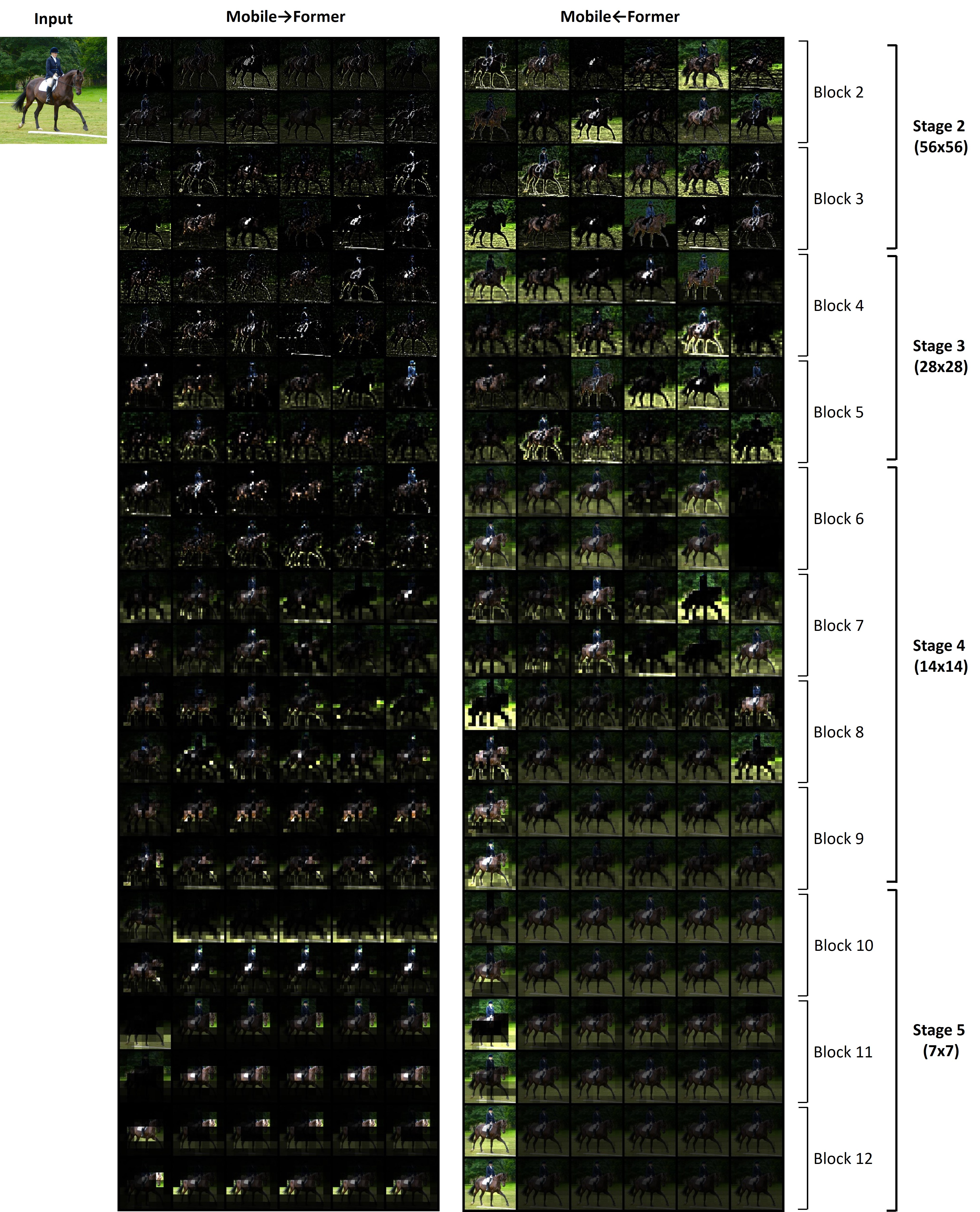}
	\end{center}
	\vspace{-4mm}
	\caption{\textbf{Visualization of the two-way cross attention}: \textit{Mobile$\rightarrow$Former} and \textit{Mobile$\leftarrow$Former}. Mobile-Former-294M is used, which includes six tokens (each corresponds to a column) and eleven Mobile-Former blocks (block 2--12) across four stages. Each block has two attention heads that are visualized in two rows. Attention in \textit{Mobile$\rightarrow$Former} (left) is normalized over pixels, showing the focused region per token. Attention in \textit{Mobile$\leftarrow$Former} (right) is normalized over tokens showing the contribution of different tokens at each pixel.}
	\label{fig:vis-all}
\end{figure*}

\vspace{2mm}
\noindent \textbf{Pattern 3 -- attention diversity across tokens diminishes:}
The attention has more diversity across tokens at low levels than high levels. As shown in \Fref{fig:vis}, each column corresponds to a token, and each row corresponds to a head in the corresponding multi-head cross attention. Note that the attention is normalized over pixels in \textit{Mobile$\rightarrow$Former} (left half), showing the focused region per token. In contrast, the attention in \textit{Mobile$\leftarrow$Former} is normalized over tokens, comparing the contribution of different tokens at each pixel.
Clearly, the six tokens at block 3 and 5 have different cross attention patterns in both \textit{Mobile$\rightarrow$Former} and \textit{Mobile$\leftarrow$Former}. Similar attention maps over tokens are clearly observed at block 8. At block 12, the last five tokens share a similar attention pattern. Note that the first token is the classification token fed into the classifier. The similar observation on token diversity has been identified in recent studies on ViT \cite{zhou2021refiner, zhou2021deepvit, touvron2021going}. 
The full visualization of two-way cross attention for all blocks is shown in \Fref{fig:vis-all}.

\end{document}